\documentclass[10pt,journal,compsoc]{IEEEtran}
\ifCLASSOPTIONcompsoc
  \usepackage[nocompress]{cite}
\else
  \usepackage{cite}
\fi
\ifCLASSINFOpdf
  \usepackage[pdftex]{graphicx}
\else
\fi
\usepackage{array}

  \usepackage[caption=false,font=footnotesize,labelfont=sf,textfont=sf]{subfig}
\usepackage{booktabs}
\usepackage{multirow}
\usepackage{upgreek}
\usepackage{bm}
\usepackage{colortbl}
\usepackage[numbers,sort&compress]{natbib}
\usepackage{hyperref}
\hypersetup{
    colorlinks=true,
    linkcolor=blue,
    filecolor=magenta,      
    urlcolor=cyan,
}

\newcommand\Tspace{\rule{0pt}{2ex}}

\begin{document}

\title{SensitiveLoss: Improving Accuracy and Fairness \\ of Face Representations with \\ Discrimination-Aware Deep Learning}

%
%

\author{Ignacio Serna, Aythami Morales, Julian Fierrez, Manuel Cebrian, Nick Obradovich, Iyad Rahwan 
\IEEEcompsocitemizethanks{\IEEEcompsocthanksitem I. Serna, A. Morales, and J. Fierrez are with the School of Engineering, Universidad Autonoma de Madrid, Spain.\protect\\
E-mail: \{ignacio.serna, aythami.morales, julian.fierrez\}@uam.es
\IEEEcompsocthanksitem M. Cebrian, N. Obradovich, I. Rahwan are with Max Planck Institute for Human Development, Berlin, Germany. \protect\\
E-mail: \{cebrian, obradovich, sekrahwan\}@mpib-berlin.mpg.de}
\thanks{Manuscript received XXX, XXX; revised XXX, XXX.}}

%
%

\markboth{Journal of \LaTeX\ Class Files,~Vol.~XX, No.~XX, XXXXXX~XXXX}%
{Shell \MakeLowercase{\textit{et al.}}: Bare Demo of IEEEtran.cls for Computer Society Journals}
%



\IEEEtitleabstractindextext{%
\begin{abstract}
We propose a discrimination-aware learning method to improve both accuracy and fairness of biased face recognition algorithms. The most popular face recognition benchmarks assume a distribution of subjects without paying much attention to their demographic attributes. In this work, we perform a comprehensive discrimination-aware experimentation of deep learning-based face recognition. We also propose a general formulation of algorithmic discrimination with application to face biometrics. The experiments include tree popular face recognition models and three public databases composed of 64,000 identities from different demographic groups characterized by gender and ethnicity. We experimentally show that learning processes based on the most used face databases have led to popular pre-trained deep face models that present a strong algorithmic discrimination. We finally propose a discrimination-aware learning method, Sensitive Loss, based on the popular triplet loss function and a sensitive triplet generator. Our approach works as an add-on to pre-trained networks and is used to improve their performance in terms of average accuracy and fairness. The method shows results comparable to state-of-the-art de-biasing networks and represents a step forward to prevent discriminatory effects by automatic systems.
\end{abstract}

}

\maketitle

\IEEEdisplaynontitleabstractindextext

%
\IEEEpeerreviewmaketitle

\IEEEraisesectionheading{\section{Introduction}\label{sec:introduction}}

\begin{figure}[!t]
\centering
\includegraphics[width=\columnwidth]{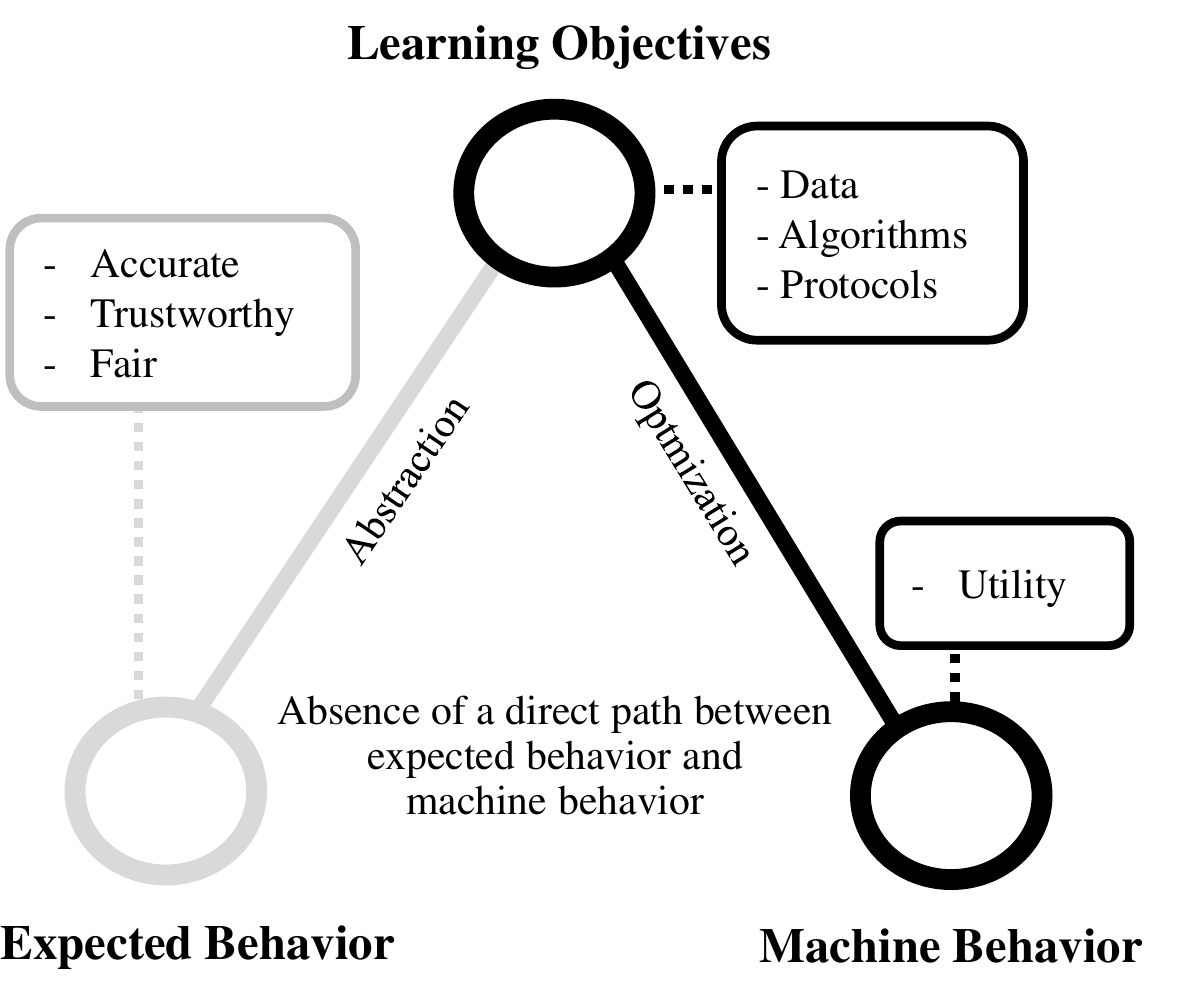}
\caption{The objective of the learning process is an abstraction of the expected behavior of an AI. There is usually no direct path between the machine expected behavior and the machine behavior, which is normally evaluated in terms of its utility. The learning objectives are usually determined by factors such as the task, data, algorithms, and experimental protocols, losing sight of key aspects in the expected behavior such as fairness. Figure inspired by the standard model proposed in \cite{russell2016AI}.}
\label{standard_model}
\end{figure}

%
\IEEEPARstart{A}{rtificial} Intelligence (AI) is developed to meet 
human needs that can be represented in the form of 
objectives. To this end, the most popular machine learning 
algorithms are designed to minimize a loss function that 
defines the cost of wrong solutions over a pool of samples. This 
is a simple but very successful scheme that has enhanced the
performance of AI in many fields such as Computer Vision, 
Speech Technologies, and Natural Language Processing. But 
this optimization of specific computable objectives may not 
lead to the behavior one may expect or desire from AI. International agencies, academia 
and industry are alerting policymakers and the public about unanticipated effects and behaviors 
of AI agents, not initially considered during the design phases \cite{rahwan2019machine}. 
In this context, aspects such as trustworthiness and fairness should be 
included as learning objectives and not taken for granted. (See Fig. \ref{standard_model}).


Machine vision in general and face recognition algorithms
in particular are good examples of recent advances in AI
\cite{Patel2018survey,marcel2020biometrics,Kumar2017book,hospedales2020deep}. The performance of automatic face recognition
has been boosted during the last decade, achieving very 
competitive accuracies in the most challenging scenarios
\cite{grother2018FRVT}. These improvements have been made possible due to 
advances in machine learning (e.g., deep learning), powerful 
computation (e.g., GPUs), and larger databases (e.g., on 
a scale of millions of images). However, the recognition 
accuracy is not the only aspect to be considered when 
designing biometric systems. Algorithms play an increasingly 
important role in the decision-making of several processes 
involving humans. So these decisions have an increasing 
impact on our lives. Thus, there is currently a growing need 
to study AI behavior in order to better understand its impact
on our society \cite{rahwan2019machine}. Face recognition systems are especially 
sensitive due to the personal information present in face
images (e.g., identity, gender, ethnicity, and age). 

The objective of a face recognition algorithm is to recognize 
when two face images belong to the same person. 
For this purpose, deep neural networks are usually trained 
to minimize a cost function over a dataset. Like many other
supervised learning processes, the training methods of these
networks consist of an iterative process where input images 
must be associated with the output labels (e.g. identities). 
This learning by imitation is highly sensitive to the 
characteristics of the dataset. The literature has demonstrated
that face recognition accuracy is affected by demographic
covariates \cite{cook2019demographic,chellapa2019face,acien2018bias, Bowyer2020face}. 
This behavior is a consequence of biases introduced into the
dataset and cost functions focused exclusively on performance
improvement. The number of published 
works pointing out the potential discriminatory effects in the 
results of face detection and recognition algorithms is large \cite{klare2012demographic,buolamwini2018GenderShades,acien2018bias,zisserman2018BlindEye,cook2019demographic,chellapa2019face,isabelle2019DemogPairs,drozdowski2020bias,Bowyer2020face}. 

In this environment, only a limited number of works
analyze how biases affect the learning process of algorithms 
dealing with personal information \cite{wang2019mitigate,gong2019debface}. 
There is a lack of understanding regarding how demographic information affects
popular and widely used pre-trained AI models beyond the performance. 

On the other hand, the right to non-discrimination is
deeply rooted in the normative framework that underlies 
various national and international regulations, and can be 
found, for example, in Article 7 of the Universal Declaration
of Human Rights and Article 14 of the European Convention
on Human Rights, among others. As evidence of these 
concerns, in April 2018 the European Parliament adopted a 
set of laws aimed at regulating the collection, storage and 
use of personal information: the General Data Protection 
Regulation (GDPR)\footnote{EU 2016/679 (General Data Protection Regulation). Available online at: https://gdpr-info.eu/}. According to paragraph 71 of GDPR,
data controllers who process sensitive data have to 
“implement appropriate technical and organizational measures …”
that “… prevent, inter alia, discriminatory effects”.

The aim of this work is to analyze face recognition models 
using a discrimination-aware perspective and to demonstrate 
that learning processes involving such discrimination-aware 
perspective can be used to train more accurate and fairer 
algorithms. The main contributions of this work are: 

\begin{itemize}
    \item A general formulation of algorithmic discrimination
    for machine learning tasks. In this work, we apply
    this formulation in the context of face recognition.
    \item A comprehensive analysis of causes and effects of
    biased learning processes including: (i) discrimination-aware 
    performance analysis based on three public datasets, with
    64K identities equally distributed across demographic 
    groups; (ii) study of deep representations and the 
    role of sensitive attributes such as gender and 
    ethnicity; (iii) complete analysis of demographic
    diversity present in some of the most popular
    face databases, and analysis of new databases 
    available to train models based on diversity.
    \item Based on our analysis of the causes and effects of biased learning algorithms, we propose an efficient discrimination-aware learning method to mitigate bias in deep face recognition models: Sensitive Loss. The method is based on the inclusion of demographic information in the popular triplet loss representation learning. Sensitive Loss incorporates fairness as a learning objective in the training process of the algorithm. The method works as an add-on to be applied over pre-trained representations and allows improving its performance and fairness without a complete re-training. We evaluate the method in three public databases showing an improvement in both overall accuracy and fairness. Our results show how to incorporate discrimination-aware learning rules to significantly reduce bias in deep learning models. 
\end{itemize}

Preliminary work in this research line was presented in 
\cite{serna2020discrimination}. Key improvements here over \cite{serna2020discrimination} include: (i) in-depth 
analysis of the state-of-the-art, including an extensive
survey of face recognition databases; (ii) inclusion of two 
new datasets in the experiments involving 40,000 new identities
and more than 1M images; and (iii) a novel discrimination-aware
learning method called Sensitive Loss.

The rest of the paper is structured as follows: Section
2 summarizes the related works. Section 3 presents our
general formulation of algorithmic discrimination. Section
4 presents the face recognition architectures used in this
work. Section 5 evaluates the causes and effects of biased
learning in face recognition algorithms. Section 6 presents 
the proposed discrimination-aware learning method. Section 
7 presents the experimental results. Finally, Section 8 
summarizes the main conclusions.

\section{Related Work}\label{Related Work}

\begin{figure*}
\centering
\includegraphics{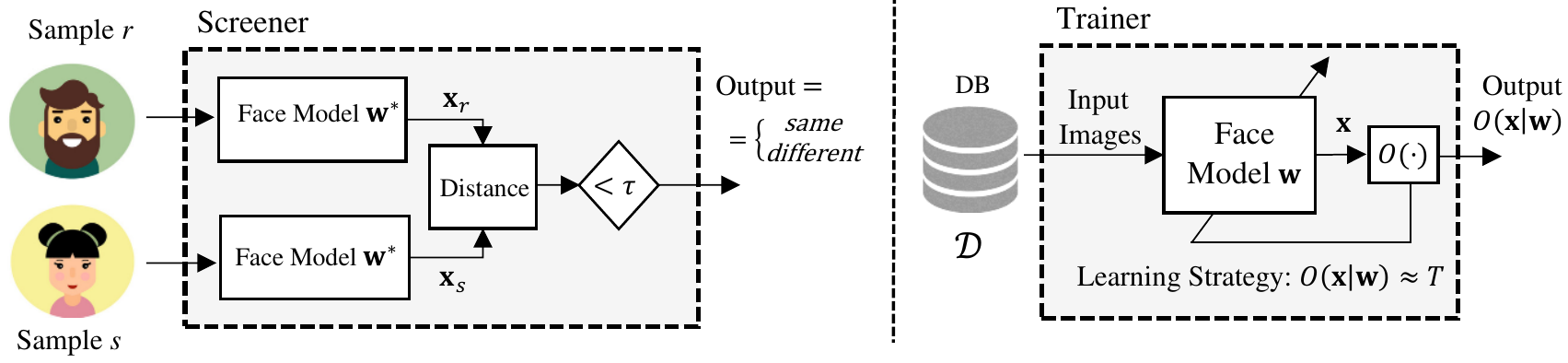}
\caption{Face recognition block diagrams. The screener is an algorithm that given two face images decides if they belong to the same person. The trainer is an algorithm that generates the best data representation for the screener.}
\label{screener_trainer}
\end{figure*}

\subsection{Face recognition: methods}

A face recognition algorithm, like other machine learning 
systems, can be divided into two different algorithms: screener 
and trainer. Both algorithms are used for a different purpose
\cite{kleinberg2019discrimination}.

The screener takes the characteristics of an individual and returns a prediction of that individual's outcome, while the trainer produces the screener itself. In our case the screener
(see Fig. \ref{screener_trainer}) is an algorithm that given
two face images generates an output associated with 
the probability that they belong to the same person. This 
probability is obtained comparing the two learned representations 
obtained from a face model defined by the parameters $\textbf{w}$. 
These parameters are trained previously based on a training 
dataset $\mathcal{D}$ (see Fig. \ref{screener_trainer}). If trained 
properly, the output of the trainer would be a model with
parameters $\textbf{w}^*$ capable of representing the input data (e.g., 
face images) in a highly discriminant feature space $\textbf{x}$.

The most popular architecture used to model face 
attributes is the Convolutional Neural Network (CNN). This 
type of network has drastically reduced the error rates of 
face recognition algorithms in the last decade \cite{ranjan2018faces} by
learning highly discriminative features from large-scale
databases.

The pre-trained models are used as an embedding extractor where $\textbf{x}$ is a $l_2$-normalised learned representation of a face image. The similarity between two face descriptors $\textbf{x}_r$ and $\textbf{x}_s$ is calculated as the Euclidean distance $||\textbf{x}_r-\textbf{x}_s||$. Two faces are assigned the same identity if their distance is smaller than a threshold $\tau$. The recognition accuracy is obtained by comparing distances between positive matches (i.e., $\textbf{x}_r$ and $\textbf{x}_s$ belong to the same person) and negative matches (i.e., $\textbf{x}_r$ and $\textbf{x}_s$ belong to different persons).

\subsection{Bias in face databases}\label{Bias in face databases}
Following the trainer-screener division, bias is rooted in the trainer. The trainer is a common algorithm that usually varies in the loss function, the optimization algorithm and in the data it uses for training. Bias is traditionally associated with the unequal representation of classes in a dataset. The history of automatic face recognition has been linked to the history of the databases used for algorithm training during the last two decades. The number of publicly available databases is high, and they allow the training of models using millions of face images. 

Table \ref{table_BD} summarizes the demographic statistics of some of the most frequently cited face databases. In order to obtain demographic statistics, gender and ethnicity classification algorithms were trained based on a ResNet-50 model \cite{he2016resnet} and 12K identities of DiveFace database (equally distributed between the six demographic groups). Models were evaluated in 20K labeled images of Celeb-A with performance over 97\%.

Each of these databases is characterized by its own biases (e.g. image quality, pose, backgrounds, and aging). In this work, we highlight the unequal representation of demographic information in very popular face recognition databases. As can be seen, the differences between ethnic groups are serious. Even though the people in Asia constitute more than 35\% of the world's population, they account for only 9\% of the content of these popular face recognition databases.

Biased databases imply a double penalty for underrepresented classes. On the one hand, models are trained according to non-representative diversity. On the other hand, accuracies are measured on privileged classes and overestimate the real performance over a diverse society.

Recently, diverse and discrimination-aware databases have been proposed in \cite{buolamwini2018GenderShades,ratha2019diversity,wang2019mitigate}. These databases are valuable resources for exploring how diversity can be used to improve face biometrics. However, some of these databases do not include identities \cite{buolamwini2018GenderShades,ratha2019diversity}, and face images cannot be matched to other images. Therefore, these databases do not allow to properly train or test face recognition algorithms. 

Note that the groups into which the databases have been divided
are heterogeneous and they include people of different 
ethnicities. We are aware of the limitations of grouping all
human ethnic origins into only three categories. According 
to studies, there are more than 5,000 ethnic groups in the 
world. Our experiments are similar to those reported in 
the literature, and include only three groups in order to 
maximize differences between classes. Automatic 
classification algorithms based on these reduced categories 
show performances of up to 98\% accuracy \cite{acien2018bias}.

\begin{itemize}
    \item[] \textbf{Algorithmic Discrimination implications}: classes $k$ are unequally represented in the most popular face databases $\mathcal{D}$. New databases and benchmarks are needed to train more diverse and heterogeneous algorithms. Evaluation over representative populations from different demographic groups is important to prevent discriminatory effects.
\end{itemize}

\begin{table*}
\centering
\renewcommand{\arraystretch}{1.2}
\caption{Demographic statistics of state-of-the-art face databases (ordered by number of images). In order to obtain demographic statistics, gender and ethnicity classification algorithms were trained based on a ResNet-50 model \cite{he2016resnet} and 12K identities of DiveFace database (equally distributed between the six demographic groups). Models were evaluated in 20K labeled images of Celeb-A with performance over 97\%. The table includes the averaged demographic statistics for the most popular face databases in the literature.}\smallskip
\label{table_BD}
    \begin{tabular}{@{}>{}p{2.5cm}>{\centering}p{1.cm}>{\centering}p{1.4cm}
                        >{\centering}p{1.5cm}>{\centering}p{1cm}>{\centering}p{1.1cm}
                        >{\centering}p{1cm}>{\centering}p{1.1cm}>{\centering}p{1cm}c@{}}
        \toprule
        & & & &\multicolumn{2}{c}{\textbf{Caucasian}}
                                            & \multicolumn{2}{c}{\textbf{African/Indian}}
                                                & \multicolumn{2}{c}{\textbf{Asian}}\\
            \cmidrule(lr){5-6}\cmidrule(lr){7-8}\cmidrule(l){9-10}
        & & & & & & & & & \\
        \multirow{-3}{*}{\textbf{Dataset [ref]}} &
        \multirow{-3}{*}{\textbf{\begin{tabular}[c]{@{}c@{}}\# \\ images\end{tabular}}} &
        \multirow{-3}{*}{\textbf{\begin{tabular}[c]{@{}c@{}}\# \\ identities\end{tabular}}} &
        \multirow{-3}{*}{\textbf{\begin{tabular}[c]{@{}c@{}}\# avg. \\ images per \\ identity\end{tabular}}}
                    & \multirow{-2}{*}{\textbf{Male}} & \multirow{-2}{*}{\textbf{Female}}
                    & \multirow{-2}{*}{\textbf{Male}} & \multirow{-2}{*}{\textbf{Female}}
                    & \multirow{-2}{*}{\textbf{Male}} & \multirow{-2}{*}{\textbf{Female}}\\ 
            \cmidrule(r){1-1}\cmidrule(lr){2-2}
            \cmidrule(lr){3-3}\cmidrule(lr){4-4}\cmidrule(lr){5-5}\cmidrule(lr){6-6}
            \cmidrule(lr){7-7}\cmidrule(lr){8-8}\cmidrule(lr){9-9}\cmidrule(l){10-10}
    
    FRVT2018 \cite{grother2019FRVT} & 27M & 12M & 2       & 48.4\% & 16.5\% & 19.9\% & 7.4\% & 1.2\% & 0.4\% \\
    MSCeleb1M \cite{2016msceleb} & 8.5M & 100K & 85    & 52.4\% & 19.2\% & 12.1\% & 3.9\% & 7.7\% & 4.5\% \\ 
    MegaFace \cite{2016Megaface}    & 4.7M & 660K & 7     & 40.0\% & 30.3\% & 6.2\% & 4.7\% & 10.6\% & 8.1\% \\ 
    VGGFace2 \cite{cao2018vgg2}    & 3.3M & 9K & 370     & 45.9\% & 30.2\% & 10.5\% & 6.3\% & 3.4\% & 3.6\% \\ 
    VGGFace \cite{parkhi2015face}     & 2.6M & 2.6K & 1K    & 43.7\% & 38.6\% & 5.8\% & 6.9\% & 2.1\% & 2.9\% \\ 
    YouTube \cite{2011youtubedb} & 621K & 1.6K & 390  & 56.9\% & 20.3\% & 7.7\% & 4.0\% & 7.9\% & 3.0\% \\ 
    CASIA \cite{2014Casiadb}   & 500K & 10.5K & 48   & 48.8\% & 33.2\% & 7.2\% & 5.7\% & 2.6\% & 2.6\% \\ 
    CelebA \cite{2015CelebA}      & 203K & 10.2K & 20   & 33.9\% & 41.5\% & 6.4\% & 8.2\% & 4.4\% & 5.5\% \\ 
    PubFig \cite{kumar2011face}      & 58K & 200 & 294     & 49.5\% & 35.5\% & 6.5\% & 5.5\% & 2.0\% & 1.0\% \\ 
    IJB-C \cite{2018IJBc}       & 21K & 3.5K & 6      & 40.3\% & 30.2\% & 11.8\% & 6.0\% & 5.4\% & 6.2\% \\ 
    UTKface \cite{utk2017age}     & 24K & - & -         & 26.2\% & 20.0\% & 21.5\% & 16.3\% & 7.1\% & 8.9\% \\ 
    LFW \cite{miller2007LFW}         & 13K & 5.7K & 2      & 58.9\% & 18.7\% & 9.6\% & 3.3\% & 7.2\% & 2.2\% \\ 
    BioSecure \cite{ortega2009BMDB}   & 2.7K & 667 & 4      & 50.1\% & 36\% & 3.1\% & 2.1\% & 4.3\% & 4.5\% \\ 
    \hline
    \textbf{Average} & & & & \textbf{46\%} & \textbf{29\%} & \textbf{10\%} & \textbf{6\%} & \textbf{5\%} & \textbf{4\%}  \\ 
    \hline \hline
    \multicolumn{10}{l}{Databases for discrimination-aware learning} \Tspace \\ \hline \Tspace
    BUPT-B \cite{wang2019mitigate} & 1.3M & 28K & 46 & \multicolumn{2}{c}{33.33\%} & \multicolumn{2}{c}{33.33\%} & \multicolumn{2}{c}{33.33\%} \\
    DiveFace \cite{SesitiveNets2019} & 125K &24K & 5 & 16.7\% & 16.7\% & 16.7\% & 16.7\% & 16.7\% & 16.7\% \\
    FairFace \cite{krkkinen2019fairface} & 100K & - & - & 25.0\% & 20.0\% & 14.4\% & 13.9\% & 13.6\% & 13.1\% \\
    RFW \cite{wang2019RFW} & 40K & 12K & 3 & \multicolumn{2}{c}{33.33\%} & \multicolumn{2}{c}{33.33\%} & \multicolumn{2}{c}{33.33\%} \\
    DemogPairs \cite{isabelle2019DemogPairs} & 10.8K & 600 & 18 & 16.7\% & 16.7\% & 16.7\% & 16.7\% & 16.7\% & 16.7\% \\ 
    \bottomrule
    \end{tabular}
\end{table*}

\subsection{Bias in face recognition}

Facial recognition systems can suffer various biases, ranging
from those derived from variables of unconstrained environments
like illumination, pose, expression and resolution of 
the face, through systematic errors such as image quality,
to demographic factors of age, gender and race \cite{chellapa2019face}.

An FBI-coauthored study \cite{klare2012demographic} tested three commercial
algorithms of supplier companies to various public organizations
in the US. In all three algorithms, African Americans
were less likely to be successfully identified —i.e., 
more likely to be falsely rejected— than other demographic 
groups. A similar decline surfaced for females compared to 
males and younger subjects compared to older subjects.

More recently, the latest NIST evaluation of commercial 
face recognition technology, the Face Recognition Vendor
Test (FRVT) Ongoing, shows that at sensitivity thresholds 
that resulted in white men being falsely matched once in $1$K,
out of a list of $167$ algorithms, all but two were more than 
twice as likely to misidentify black women, some reaching
$40$ times more \cite{grother2019FRVT}. The number of academic
studies analyzing fairness of face recognition algorithms has 
grown during last years \cite{drozdowski2020bias}.

There are other published studies analyzing face recognition
performance over demographic groups, but  \cite{klare2012demographic} and
\cite{grother2019FRVT}, are probably the most systematic, comprehensive,
thorough, and up-to-date.

\subsection{De-biasing face recognition}

There are attempts to eliminate bias in face recognition, as
in \cite{zisserman2018BlindEye}, with so-called unlearning, which improves
the results, but at the cost of losing recognition accuracy. Das et al. 
proposed a Multi-Task CNN that also managed to improve 
performance across subgroups of gender, race, and age \cite{das2018mitigatebias}. 
Finally, in \cite{SesitiveNets2019} an extension of the triplet loss function
is developed to remove sensitive information in feature
embeddings, without losing performance in the main task.

In \cite{wang2019mitigate}, researchers proposed a race-balanced reinforcement learning
network to adaptively find appropriate margins losses for
the different demographic groups. Their model significantly
reduced the performance difference obtained between demographic
groups. \cite{gong2019debface} with an adversarial network, 
disentangles feature representation of gender, age, race and face 
recognition and minimizes their correlation. Both methods
\cite{wang2019mitigate,gong2019debface} were applied to train de-biasing deep 
architectures for face recognition from scratch.

\section{Formulating Algorithmic Discrimination}
Discrimination is defined by the Cambridge Dictionary as 
treating a person or particular group of people differently,
especially in a worse way than the way in which you treat
other people, because of their skin color, sex, sexuality, etc.

For the purpose of studying discrimination in artificial 
intelligence at large, we now formulate mathematically 
\emph{Algorithmic Discrimination} based on the above dictionary 
definition. Even though ideas similar to those included in 
our formulation can be found elsewhere \cite{calders2010discrimination,raji2019bias}, we didn't 
find this kind of formulation in related works. We hope 
that the formalization of these concepts can be beneficial
in fostering further research and discussion on this hot topic.

Let’s begin with notation and preliminary definitions. 
Assume $\textbf{x}_s^i$ is a learned representation of individual $i$ (out of
$I$ different individuals) corresponding to an input image $\textbf{I}^i_s$
($s=1,\ldots,S$ samples per individual). That representation $\textbf{x}$ 
is assumed to be useful for task $T$, e.g., face authentication or 
emotion recognition. That representation $\textbf{x}$ is generated 
from the input image $\textbf{I}$ using an artificial intelligence 
approach with parameters $\textbf{w}$. We also assume that there is 
a goodness criterion $G$ in that task that maximizes some 
real-valued performance function $f$ in a given dataset $\mathcal{D}$
(collection of multiple images) in the form:

\begin{equation}
\label{eqn:goodnes_criterion}
    \textit{G}(\mathcal{D}) = \max_{\textbf{w}}\textit{f}(\mathcal{D},\textbf{w})
\end{equation}

The most popular form of the previous expression minimizes 
a loss function $\mathcal{L}$ over a set of training images 
$\mathcal{D}$ in the form:

\begin{equation}
\label{eqn:learning_strategy}
    \textbf{w}^*=\arg\min_{\textbf{w}}{\sum_{\textbf{I}_s^j\in \mathcal{D}}\mathcal{L}(\textit{O}(\textbf{I}_s^j|\textbf{w}),T^j_s)} 
\end{equation}

\noindent where \textit{O} is the output of the learning algorithm that we
seek to bring closer to the target function (or groundtruth)
\textit{T} defined by the task at hand. On the other hand, the \textit{I}
individuals can be classified according to \textit{D} demographic 
criteria $\textit{C}_d$, with $d = 1,..., D$, which can be the source
for discrimination, e.g., $\textit{C}_1 = \textit{Gender} = \{\textit{Male, Female}\}$
(the demographic criterion $\textit{Gender}$ has two classes in
this example). The particular class $k=1,...,K$ for a given
demographic criterion $d$ and a given sample is noted as 
$\textit{C}_d (\textbf{x}_s^i)$, e.g., $\textit{C}_1 (\textbf{x}_s^i)=\textit{Male}$. We assume that all classes are 
well represented in dataset $\mathcal{D}$, i.e., the number of samples
for each class in all criteria in $\mathcal{D}$ is significant. $\mathcal{D}_d^k \in \mathcal{D}$ 
represents all the samples corresponding to class $k$ of
demographic criterion $d$.

Finally, \textbf{our definition of Algorithmic Discrimination}:
\begin{itemize}
    \item[] An algorithm discriminates the group of people 
represented with class $k$ (e.g., \textit{Female}) when
performing the task \textit{T} (e.g., face verification), 
if the goodness \textit{G} in that task
when considering the full set of data $\mathcal{D}$ (including
multiple samples from multiple individuals), is significantly 
larger than the goodness $\textit{G}(\mathcal{D}_d^k)$ in 
the subset of data corresponding to class $k$ of the 
demographic criterion $d$.
\end{itemize}

The representation $\textbf{x}$ and the model parameters $\textbf{w}$ will
typically be real-valued vectors, but they can be any set of
features combining real and discrete values. Note that the
previous formulation can be easily extended to the case
of varying number of samples $S_i$  for different subjects, 
which is a usual case; or to classes \textit{K} that are not
disjoint. Note also that the previous formulation is based on 
average performances over groups of individuals. In many 
artificial intelligence tasks it is common to have different 
performance between specific individuals due to various 
reasons, e.g., specific users who were not sensed properly
\cite{fierrez2011quality}, even in the case of algorithms that, on average, may 
have similar performance for the different classes that are
the source of discrimination. Therefore, in our formulation
and definition of Algorithmic Discrimination we opted to 
use average performances in demographic groups.

Other related works are now starting to investigate 
discrimination effects in AI with user-specific methods, e.g.
\cite{pentland2020fair,varshney2020AIfairness}, but they are still lacking a mathematical framework with 
clear definitions of User-specific Algorithmic Discrimination 
(U-AD), in comparison to our defined Group-based Algorithmic
Discrimination (G-AD). We will study and augment our
framework with an analysis of U-AD in future work.

\section{Proposed Approach: Sensitive Loss}

Models trained and evaluated over privileged demographic groups may fail to generalize when the model is evaluated over groups other than the privileged one. This is a behavior caused by the wrong assumption of homogeneity in face characteristics of the world population. In this work we propose to reduce the bias in face recognition models incorporating a discrimination-aware learning process.

The methods proposed in this work to reduce bias are based on two strategies:
\begin{enumerate}
    \item[i)] Use of balanced and heterogeneous data to train and evaluate the models. The literature showed that training with balanced dataset does not guarantee bias-free results \cite{wang2019RFW,wang2019mitigate,klare2012demographic} but can partially reduce it.
    \item[ii)]  A modified loss function (Sensitive Loss) that incorporates demographic information to guide the learning process into a more inclusive feature space. The development of new cost functions capable of incorporating discrimination-aware elements into de training process is another way to reduce bias. Our approach is based on the popular triplet loss function and it can be applied to pre-trained models without needing the full re-training of the network.
\end{enumerate}

\subsection{Discrimination-aware learning into triplet loss}\label{SensitiveLoss}

\begin{figure*}
\centering
\includegraphics[width=0.9\textwidth]{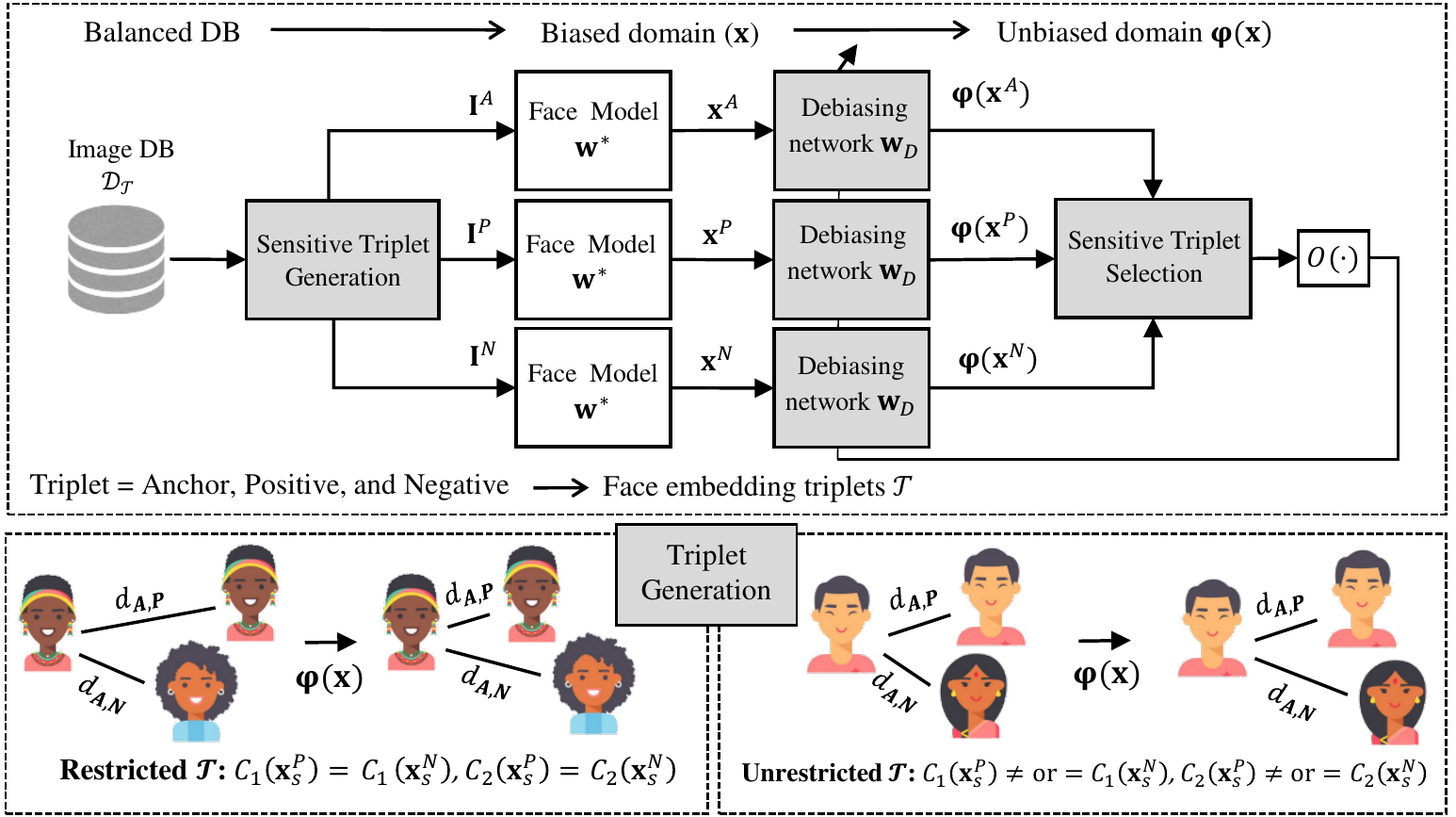}
\caption{(Up) Block diagram of the domain adaptation learning process that allows us to generate an unbiased representation $\bm{\upvarphi}(\textbf{x})$ from a biased representation $\textbf{x}$. A Balanced Dataset $\mathcal{D}_\mathcal{T}$ is preferable as input to train Sensitive Loss for selecting the triplets $\mathcal{T}$. This $\mathcal{D}_\mathcal{T}$ can be different or a subset of the (generally Unbalanced) Dataset $\mathcal{D}$ used for training the biased model $\textbf{w}^*$ that appears in Eq. \ref{eqn:goodnes_criterion}. (Down) Discrimination-aware generation of triplets given an underrepresented (unfavored) demographic group: the representation $\bm{\upvarphi}(\textbf{x})$ increases the distance $d$ between Anchor and Negative samples while reducing the distance between Anchor and Positive, trying in this way to improve the performance of the unfavored group.}
\label{triplets}
\end{figure*}

Triplet loss was proposed as a distance metric in the context of nearest neighbor classification \cite{weinberger2006Triplet} and adapted to improve the performance of face descriptors in verification algorithms \cite{schroff2015facenet,parkhi2015face}. In this work we propose to incorporate demographic data to generate discrimination-aware triplets to train a new representation that mitigates biased learning. 

Assume that an image is represented by an embedding 
descriptor $\textbf{x}_s^i$ obtained by a pre-trained model
(see Section 3 for notation). That image corresponds to 
the demographic group $C_d{(\textbf{x}_s^i)}$. A triplet is composed of 
three different images of two different people: Anchor ($A$)
and Positive ($P$) are different images of the same person, 
and Negative ($N$) is an image of a different person. The 
Anchor and Positive share the same demographic labels,
$C_d{(\textbf{x}_s^A)}=C_d{(\textbf{x}_s^P)}$ but these labels may differ for the 
Negative sample $C_d{(\textbf{x}_s^N)}$. The transformation $\bm{\upvarphi}(\textbf{x})$ represented by 
parameters $\textbf{w}_D$ ($D$ for De-biasing) is trained to minimize
the loss function:

\begin{equation}
\label{eqn:triplet_loss}
    \min_{\textbf{w}_D}{\sum_{\textbf{x}_s\in \mathcal{T}}(||\bm{\upvarphi}(\textbf{x}_s^A)-\bm{\upvarphi}(\textbf{x}_s^N)||^{2}-||\bm{\upvarphi}(\textbf{x}_s^A)-\bm{\upvarphi}(\textbf{x}_s^P)||^{2}+\Delta)} 
\end{equation}

\noindent where $||\cdot||$ is the Euclidean Distance, $\Delta$ is a margin between genuine and impostor distances, and $\mathcal{T}$ is a set of triplets generated by an online sensitive triplet generator that guides the learning process (see details in Section \ref{sensitive_triplets}). The effects of biased training include a representation that fails to model properly the distance between faces from different people ($||\textbf{x}_s^A-\textbf{x}_s^N||$) belonging to the same minority demographic groups (e.g. $C_d{(\textbf{x}_s^A)}=C_d{(\textbf{x}_s^N)} = \textit{Asian Female}$). The proposed triplet loss function considers both genuine and impostor comparisons and also allows to introduce demographic-aware information. In order to guide the learning process in that discrimination-aware spirit, triplets from demographic groups with worst performances are prioritized in the online sensitive triplet generator (e.g. for $\textit{Asian Females}$). Fig. \ref{triplets} shows the block diagram of the learning algorithm.

\subsection{Sensitive Loss: sensitive triplets} \label{sensitive_triplets}

Inspired in the semi-hard selection proposed in \cite{schroff2015facenet,parkhi2015face}, we propose an online selection of triplets that prioritizes the triplets from demographic groups with lower performances (see Fig. \ref{triplets}). On the one hand, triplets within the same demographic group improve the ability to discriminate between samples with similar anthropometric characteristics (e.g. reducing the false acceptance rate in $\textit{Asian Females}$). On the other hand, heterogeneous triplets (i.e. triplets involving different demographic groups) improve the generalization capacity of the model (i.e. the overall accuracy). 

During the training process we distinguish between generation and selection of triplets:

\begin{itemize}
    \item \textit{Triplet Generation}: this is where the triplets are formed and joined to compose a training batch. In our experiments, each batch is generated randomly with images from $300$ different identities equally distributed among the different demographic groups ($900$ images in total). We propose two types of triplets generation (see Fig. \ref{triplets}):
    \begin{itemize}
        \item \textbf{Unrestricted (U)}: the generator allows triplets with mixed demographic groups (i.e. $\textit{C}_d (\textbf{x}_s^A)=\textit{C}_d(\textbf{x}_s^N)$ or $\textit{C}_d (\textbf{x}_s^A) \neq \textit{C}_d(\textbf{x}_s^N)$), but identities are equally distributed, that is, there are the same number of identities for each demographic group. Thus, with 300 identities, around 135K triplets are generated (from which the semi-hard ones will be selected).
        \item \textbf{Restricted (R)}: the generator does not allow triplets with mixed demographic groups (i.e. $\textit{C}_d (\textbf{x}_s^P)=\textit{C}_d(\textbf{x}_s^N)$) and identities are equally distributed. Thus, with 300 identities, more than 22K triplets are generated (from which the semi-hard ones will be selected).
    \end{itemize}
    \item \textit{Triplet Selection}: Triplet selection is done online during the training process for efficiency. Among all the triplets in the generated batches, the online selection chooses those for which: ${0}<||\textbf{x}_s^A-\textbf{x}_s^N||^{2}-||\textbf{x}_s^A-\textbf{x}_s^P||^{2}<\Delta$. These are semi-hard triplets and are crucial to have an adequate convergence and not to lead to bad local minima \cite{schroff2015facenet}. If a demographic group is not well modeled by the network (both in terms of genuine or impostor comparisons), more triplets from this group are likely to be included in the online selection. This selection is purely guided by performance over each demographic group and could change for each batch depending on model deficiencies.
\end{itemize}

We chose triplet loss as the basis for Sensitive Loss because it allows us to incorporate the demographic-aware learning in a natural way. The process is data driven and does not require a large number of images per identity (e.g. while softmax requires a large number of samples per identity we only use $3$ images per identity). Another advantage is that it is not necessary to train the entire network, and triplet loss can be applied as a domain adaptation technique. In our case, we trained the model to move from a biased domain $\textbf{x}$ to an unbiased domain $\bm{\upvarphi}(\textbf{x})$. Our results demonstrate that biased representations $\textbf{x}$ that exhibit clear performance differences contain the information necessary to reduce such differences. In other words, bias can be at least partially corrected from representations obtained from pre-trained networks, and new models trained from scratch are not necessary. Similar strategies might be applied to other loss functions.

\section{Evaluation Procedure}

\subsection{Databases for discrimination-aware learning}\label{subsection:Databases}

\emph{DiveFace} \cite{SesitiveNets2019} contains annotations equally distributed among 
six classes related to gender and ethnicity. There are 24K 
identities (4K per class) and 3 images per identity for a 
total number of images equal to 72K. Users are grouped 
according to their gender (male or female) and three 
categories related with ethnic physical characteristics: \textit{Caucasian}: people with ancestral origins in Europe, 
North-America, and Latin-America (with European origin).
\textit{African/Indian}: people with ancestral origins in 
Sub-Saharan Africa, India, Bangladesh, Bhutan, among others. \textit{Asian}: people with ancestral origin in Japan,
China, Korea, and other countries in that region.

\emph{Races Face in the Wild (RFW)} \cite{wang2019RFW}
is divided into four demographic classes: Caucasian, Indian, Asian and African. Each class has about 10K images of 3K individuals. There are no major differences in pose, age and gender distribution between Caucasian, Asian and Indian groups. The African set has smaller age difference than the others, and while in the other groups women represent about 35\%, in Africans they represent less than 10\%.

\emph{BUPT-Balancedface (BUPT-B)} \cite{wang2019mitigate} contains 1.3M images from 28K celebrities obtained from MS-Celeb-1M \cite{2016msceleb}. Divided into 4 demographic groups, it is roughly balanced by race with 7K subjects per race: Caucasian, Indian, Asian, and African; with 326K, 325K, 275K and 324K images respectively. No gender data is available for this dataset.

\subsection{Deep face recognition models}

\emph{VGG-Face \cite{parkhi2015face}:} Model based on the VGG-Very-Deep-16 CNN traditional architecture with 138M parameters. We used a pre-trained model\footnote{\label{note1}Available on https://github.com/rcmalli/keras-vggface} trained with the VGGFace2 dataset according to the details provided in \cite{cao2018vgg2}. The VGG models were developed by the Visual Geometry Group (VGG) at the University of Oxford for face recognition and demonstrated on benchmark computer vision datasets \cite{parkhi2015face}.

\emph{ResNet-50 \cite{he2016resnet}:} ResNet-50 is a CNN model with 25M parameters initially proposed for general purpose image recognition tasks \cite{he2016resnet}. Combines convolutional neural networks with residual connections to allow information to skip layers and improve gradient flow. These models have been tested on competitive evaluations 
and public benchmarks \cite{parkhi2015face,cao2018vgg2}. The model\textsuperscript{\ref{note1}} we used has been trained with the VGGFace2 dataset.

\emph{ArcFace \cite{deng2019arcface}:} With a ResNet architecture and $64$M parameters, ArcFace obtains state-of-the-art results on multiple datasets (e.g. 99.80\% accuracy on LFW\cite{miller2007LFW}). We used a publicly available\footnote{https://github.com/deepinsight/insightface} pre-trained ArcFace model trained on MS-Celeb-1M \cite{2016msceleb}.

\subsection{Implementation details}

The proposed de-biasing method Sensitive Loss does not require retraining the entire pre-trained model (see Fig. \ref{triplets}). The sensitive triplets are used to train a dense layer with the following characteristics: number of units equal to the size of the pre-trained representation $\textbf{x}$ ($4$,$096$, $2$,$048$ and $512$ units for VGG-Face, ResNet-50 and ArcFace respectively), dropout (of 0.5 for VGG-Face and Resnet-50 and 0.05 for ArcFace), linear activation, random initialization, and $L_{2}$ normalization. This layer, relatively easy to train (10 epochs and Adam optimizer), will be used to generate the new representation $\bm{\upvarphi}(\textbf{x})$.

The experiments are carried out with $k$-fold cross-validation across users and three images per identity (therefore $3$ genuine and $3 \times (\textrm{users}-1)$ impostor combinations per identity), with 5 folds. Thus, the three databases are divided into a training set ($80\%$) and a test set ($20\%$) in every fold. Resulting in a total of $192$K genuine comparisons (DiveFace $=72$K, RFW $=36$K y BUPT $= 84$K) and $98$M impostor comparisons (DiveFace $=28.7$M, RFW $10.8$M y BUPT $= 58.7$M).

Note that ArcFace is only evaluated on DiveFace since that model was trained with MS1M \cite{2016msceleb}, which overlaps with RFW and BUPT-B datasets (both databases obtained from it). Before applying the face recognition models, we cropped the face 
images using the algorithms proposed in \cite{zhang2016detection,deng2020retina}.

\section{Experiments}


\subsection{Demographic bias in learned representations}

We applied a popular data visualization algorithm to better understand the importance of ethnic features in the embedding space generated by deep models. t-SNE is an algorithm to visualize high-dimensional data. This algorithm minimizes the Kullback-Leibler divergence between the joint probabilities of the low-dimensional embedding and the high-dimensional data.

\begin{figure}
\begin{center}
\includegraphics[width=\columnwidth]{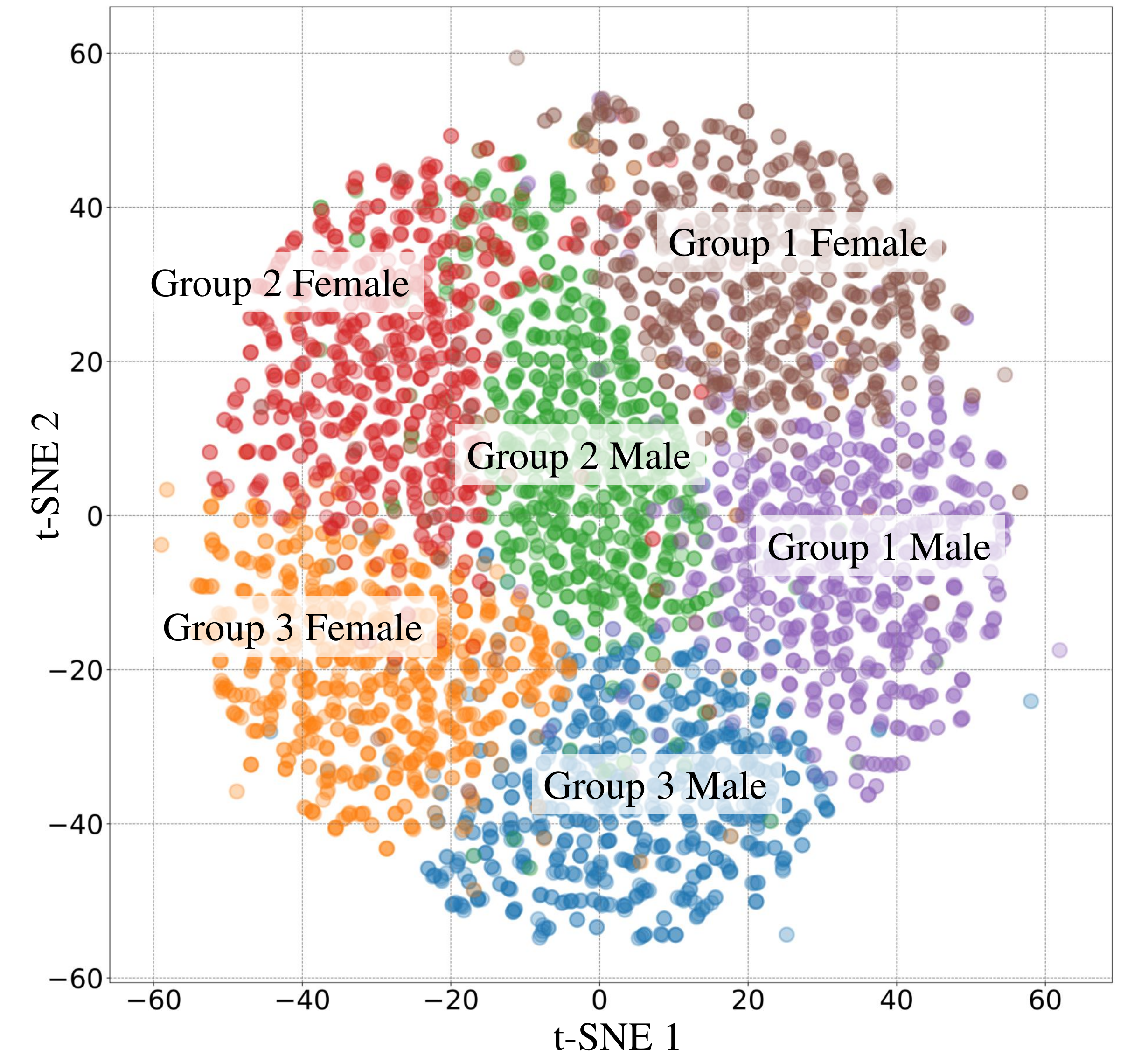}
        \label{fig:tsne}
\caption{Projections of the ResNet-50 embeddings into the 2D space generated with t-SNE. (This is a colored image.)}
\end{center}
\end{figure}

Fig. \ref{fig:tsne} shows the projection of each face into a 2D space generated from ResNet-50 embeddings and the t-SNE algorithm. This t-SNE projection is unsupervised and just uses as input the face embeddings without any labels. After running t-SNE, we have colored each projected point according to its ethnic attribute. As we can see, the consequent face representation results in three clusters highly correlated with the ethnicity attributes. Note that ResNet-50 has been trained for face recognition, not ethnicity detection. However, the gender and ethnicity information is highly embedded in the feature space and the unsupervised t-SNE algorithm reveals the presence of this information.

We have used the t-SNE algorithm also on the embeddings of our method, and the result is the same as in Fig.\ref{fig:tsne}. This means that ethnic information is still present. We do not propose to remove racial information, this information is key to recognition and it isnot our goal to remove it, in fact removing it decreases performance, as shown in  \cite{gong2019debface}.

On a different front, CNNs are composed of a large number of stacked filters. These filters are trained to extract the richest information for a pre-defined task (e.g. face recognitionin). Since face recognition models are trained to identify individuals, it is reasonable to think that the response of the models may vary slightly from one person to another. In order to visualize the response of the model to different faces, we consider the specific Class Activation MAP (CAM) proposed in \cite{selvaraju2017grad}, named Grad-CAM. This visualization technique uses the gradients of a target flowing into the selected convolutional layer to produce a coarse localization map. The resulting heatmap highlights the activated regions in the image for the selected target (e.g. an individual identity in our case).

\begin{figure}
\begin{center}
\includegraphics[width=\columnwidth]{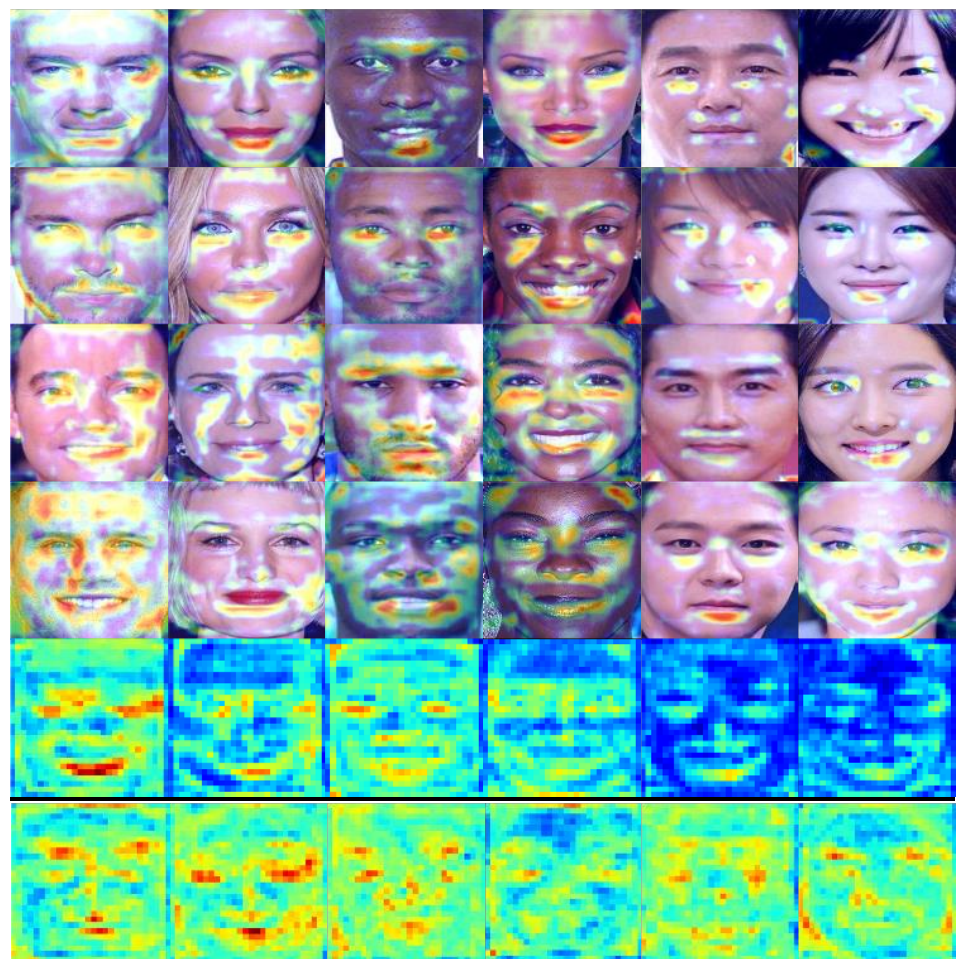}
        \label{fig:activations}
\caption{Examples of the six classes available in the DiveFace database (different columns). Rows 5 and 6 show the averaged Class Activation MAP (first filter of the third convolutional block of ResNet-50) with and without our method obtained from 20 random face images from each of the classes. Rows 1-4 show Class Activation MAPs for each of the face images. Maximum and minimum activations are represented by red and blue colors respectively. (This is a colored image.) }
\end{center}
\end{figure}

Fig. \ref{fig:activations} represents the heatmaps obtained in the first filter of the third convolutional block of ResNet-50 for faces from the six demographic groups included in DiveFace. Each column corresponds to a demographic group. The first rows contain face images with their heatmap superimposed. The last two rows represent the heatmaps obtained in the same ResNet-50 filter without and with our method after averaging results from 120 different individuals. For a better visualization the 120 images chosen are all frontal. We only averaged a small group of individuals because if we did it with the whole dataset nothing would be seen, since the images vary widely in pose and morphology.

The activation maps show clear differences between ethnic groups with the highest activation for caucasians and the lowest for asians. These differences suggest that features extracted by the model are, at least, partially affected by the ethnic attributes. However, with our method (last row) the activations are more homogeneous across demographic groups. This homogeneous activation suggests a better representation across the different ethnic groups. Recent work has shown that there is a correlation between high activations and performance in CNN's architectures \cite{bau2020understanding}.  The activation maps obtained with the VGG-Face and ArcFace models are similar to those of ResNet-50. 

These two experiments illustrate the presence and importance of ethnic attributes in the feature space generated by face deep models. 

\begin{itemize}
    \item[] \textbf{Algorithmic Discrimination implications}: popular deep models trained for task \textit{T} on biased databases (i.e., unequally represented classes $k$ for a given demographic criterion $d$ such as gender) result in feature spaces (corresponding to the solution $\textbf{w}^*$ of the Eq. \ref{eqn:learning_strategy}) that introduce strong differentiation between classes $k$. This differentiation affects the representation $\textbf{x}$ and enables classifying between classes $k$ using $\textbf{x}$, even though \textbf{x} was trained for solving a different task ${T}$.
\end{itemize}

\begin{table*}
\centering
\caption{Face verification Performance (Equal Error Rate EER in \%) on the face datasets described in Section \ref{subsection:Databases} for matchers VGG-Face, ResNet-50 and ArcFace without and with our de-biasing Sensitive Loss module (U = Unrestricted Triplet Generation; R = Restricted Triplet Generation). Also shown: Average EER across demographic groups and Standard deviation (lower means fairer).}\smallskip

\begin{tabular}{>{\raggedright}p{1.7cm}>{\centering}p{1.2cm}>{\centering}p{1.2cm}>{\centering}p{1.2cm}
              >{\centering}p{1.2cm}>{\centering}p{1.2cm}>{\centering}p{1.2cm}>{\centering}p{1.9cm}>{\centering}p{1.8cm}}
    \toprule
    
    \multirow{3}{*}{\textbf{Model}} & \multicolumn{8}{c}{\textbf{DiveFace}}
    \tabularnewline
    \cmidrule(lr){2-9}
    & \multicolumn{2}{c}{\textbf{Caucasian}}    & \multicolumn{2}{c}{\textbf{Indian/African}}    & \multicolumn{2}{c}{\textbf{Asian}}    & \textbf{}    &
    \tabularnewline
                
    \cmidrule(lr){2-3}\cmidrule(lr){4-5}\cmidrule(l){6-7}
        
    & {\textbf{Male}} & {{\textbf{Female}}} & {\textbf{Male}} & {{\textbf{Female}}} & {\textbf{Male}} & {{\textbf{Female}}}   & \textbf{Avg} & \textbf{Std}
    \tabularnewline
                            
    \cmidrule(r){1-1}\cmidrule(lr){2-2}\cmidrule(lr){3-3}\cmidrule(lr){4-4}\cmidrule(lr){5-5}\cmidrule(lr){6-6}\cmidrule(lr){7-7}\cmidrule(lr){8-8}\cmidrule(l){9-9}
        
    {\footnotesize\textrm{VGG-Face}} & 1.62 & 1.76 & 2.06 & 2.33 & 2.53 & 3.15 & \textbf{2.24} & \textbf{0.51}
    \tabularnewline
    {\footnotesize\textrm{VGG-Face-U}} & 1.84 & 1.98 & 1.63 & 1.77 & 1.38 & 1.44 & 1.67 ($\downarrow$25\%)      & 0.21 ($\downarrow$58\%)
    \tabularnewline
    {\footnotesize\textrm{VGG-Face-R}} & 1.80 & 1.97 & 1.65 & 1.77 & 1.42 & 1.42 & 1.67 ($\downarrow$25\%)      & 0.20 ($\downarrow$61\%)
    \tabularnewline
    \midrule
    {\footnotesize\textrm{ResNet-50}} & 0.63 & 0.73 & 0.88 & 1.41 & 0.99 & 1.26 &\textbf{0.98} &\textbf{0.28}
    \tabularnewline
    {\footnotesize\textrm{ResNet-50-U}} & 0.84 & 0.90 & 0.74 & 1.21 & 0.58 & 0.60 & 0.81 ($\downarrow$17\%) & 0.21 ($\downarrow$24\%)
    \tabularnewline
    {\footnotesize\textrm{ResNet-50-R}} & 0.90 & 0.93 & 0.78 & 1.22 & 0.61 & 0.62 & 0.84 ($\downarrow$14\%) & 0.21 ($\downarrow$25\%)
    \tabularnewline
    \midrule
    {\footnotesize\textrm{ArcFace}} & 0.79 & 0.85 & 1.11 & 1.98 & 1.34 & 1.27 &\textbf{1.22} &\textbf{0.39}
    \tabularnewline
    {\footnotesize\textrm{ArcFace-U}} & 0.71 & 0.67 & 1.08 & 1.79 & 1.24 & 1.17 & 1.11 ($\downarrow$9\%) & 0.37 ($\downarrow$5\%)
    \tabularnewline
    {\footnotesize\textrm{ArcFace-R}} & 0.69 & 0.65 & 0.96 & 1.88 & 1.22 & 1.19 & 1.10 ($\downarrow$10\%) & 0.41 ($\uparrow$5\%)
    \tabularnewline
    
\end{tabular}

\begin{tabular}{>{\raggedright}p{2cm}>{\centering}p{1.7cm}>{\centering}p{1.7cm}>{\centering}p{1.7cm}
              >{\centering}p{1.7cm}>{\centering}p{2.3cm}>{\centering}p{2.3cm}}
    \toprule
    
    \multirow{3}{*}{\textbf{Model}} & \multicolumn{6}{c}{\textbf{RFW}}
    \tabularnewline
    
    \cmidrule(lr){2-7}
                                                
    \multirow{3}{*}{}    & \multirow{2}{*}{{\textbf{Caucasian}}}    & \multirow{2}{*}{\textbf{Indian}}    & \multirow{2}{*}{{\textbf{African}}}    & \multirow{2}{*}{{\textbf{Asian}}}    &  &
    \tabularnewline

    & & & & & \textbf{Avg} & \textbf{Std}
    \tabularnewline
    
    \cmidrule(r){1-1}\cmidrule(lr){2-2}\cmidrule(lr){3-3}\cmidrule(lr){4-4}\cmidrule(lr){5-5}\cmidrule(lr){6-6}\cmidrule(lr){7-7}
    
    {\footnotesize\textrm{VGG-Face}} & 8.22 & 10.38 & 17.24 & 13.67 & \textbf{12.38}     & \textbf{3.41}
    \tabularnewline
    {\footnotesize\textrm{VGG-Face-U}} & 7.34 & 7.78 & 13.09 & 9.47 & 9.42 ($\downarrow$24\%)      & 2.27 ($\downarrow$34\%)
    \tabularnewline
    {\footnotesize\textrm{VGG-Face-R}} & 7.26 & 7.75 & 12.79 & 9.05 & 9.21 ($\downarrow$26\%)      & 2.17 ($\downarrow$36\%)
    \tabularnewline
    \midrule
    {\footnotesize\textrm{ResNet-50}} & 3.62 & 4.72 & 5.75 & 5.96 & \textbf{5.01} & \textbf{0.93}
    \tabularnewline
    {\footnotesize\textrm{ResNet-50-U}} & 3.02 & 3.29 & 3.99 & 3.83 & 3.53 ($\downarrow$30\%) & 0.40 ($\downarrow$58\%)
    \tabularnewline
    {\footnotesize\textrm{ResNet-50-R}} & 3.02 & 3.22 & 4.06 & 3.92 & 3.56 ($\downarrow$29\%) & 0.44 ($\downarrow$53\%)
    \tabularnewline

\end{tabular}

\begin{tabular}{>{\raggedright}p{2cm} >{\centering}p{1.7cm} >{\centering}p{1.7cm} >{\centering}p{1.7cm}
              >{\centering}p{1.7cm} >{\centering}p{2.3cm} >{\centering}p{2.3cm}}
    \toprule
    
    & \multicolumn{6}{c}{\textbf{BUPT-Balanceface}} 
    \tabularnewline
    
    \cmidrule(lr){2-7}
    
    {\footnotesize\textrm{VGG-Face}} & 7.18 & 7.44& 9.78 & 12.56  &\textbf{9.24} &\textbf{2.17}
    \tabularnewline
    {\footnotesize\textrm{VGG-Face-U}} & 6.49 & 4.97 & 7.73 & 8.29 & 6.87 ($\downarrow$26\%)      & 1.28 ($\downarrow$41\%)
    \tabularnewline
    {\footnotesize\textrm{VGG-Face-R}} & 6.48 & 5.03 & 7.68 & 8.20 & 6.85 ($\downarrow$26\%)      & 1.22 ($\downarrow$44\%)
    \tabularnewline
    \midrule
    {\footnotesize\textrm{ResNet-50}} & 3.24 & 2.65 & 3.80 & 5.56 &\textbf{3.82} &\textbf{1.09}
    \tabularnewline
    {\footnotesize\textrm{ResNet-50-U}} & 2.62 & 1.69 & 2.72 & 3.19 & 2.56 ($\downarrow$33\%) & 0.54 ($\downarrow$50\%)
    \tabularnewline
    {\footnotesize\textrm{ResNet-50-R}} & 2.62 & 1.72 & 2.77 & 3.12 & 2.56 ($\downarrow$32\%) & 0.52 ($\downarrow$52\%)
    \tabularnewline

\end{tabular}

\label{tabla_EER}
\end{table*}

\subsection{Performance of Sensitive Loss}

Table \ref{tabla_EER} shows the performance (Equal Error Rate EER in \%) for each demographic group as well as the average EER on the DiveFace, RFW and BUPT test sets for the baseline models (VGG-Face, ResNet-50 and ArcFace), and the Sensitive Loss methods described in Section \ref{SensitiveLoss} (Unrestricted and Restricted). In order to measure the fairness, Table \ref{tabla_EER} includes the Standard deviation of the EER across demographic groups (Std). Theses measures were proposed in \cite{wang2019mitigate,gong2019debface} to analyze the performance of de-biasing algorithms.

If we focus on the results obtained by the Baseline systems (denoted as VGG-Face, ResNet-50 and ArcFace), the different performances obtained for similar demographic groups in the three databases are caused by the different characteristics of each database (e.g. the African set has a smaller age difference than the others in RFW). The results reported in Table \ref{tabla_EER} exhibit large gaps between the performances obtained by the different demographic groups, suggesting that both gender and ethnicity significantly affect the performance of biased models. These effects are particularly high for ethnicity, with a very large degradation in performance for the class less represented in the training data. For DiveFace, this degradation produces a relative increment of the Equal Error Rate (EER) of $94\%$, $124\%$ and $150\%$ for VGG-Face, ResNet-50 and ArcFAce, respectively, with regard to the best class (\textit{Caucasian Male}). For RFW and BUPT-Balanceface the differences between demographic groups are similar but not so large, because the distinction between the demographic groups is only of ethnic origin and not of sex.

These differences are important as they mark the percentage of faces successfully matched and faces incorrectly matched for a certain threshold. These results indicate that ethnicity can greatly affect the chances of being mismatched (false positives).

Concerning the triplet generation method (Unrestricted or Resticted, see Section \ref{sensitive_triplets}), both methods show competitive performances with similar improvements over the baseline approaches. The higher number of triplets generated by the Unrestricted method (about $6$ times more) does not show clear improvements compared to the Restricted method. We can see that the biggest improvements are achieved for VGG-Face, and that ArcFace barely improves (in fact ArcFace-R worsens its std by 5\%). This is due to the fact that the size of the embedding obtained from VGG-Face is 8 times larger than in ArcFace, a model whose performance is already highly optimized, and therefore the margin of improvement in VGG-Face and ResNet-50 is much greater.

One would think that ResNet-50 is not part of the state of the art. Yet, in Table \ref{tabla_EER} you can see that ResNet-50 has a better average accuracy (lower average EER) than ArcFace in the DiveFace database. Normally performance evaluations are done on unbalanced datasets (See Table \ref{Bias in face databases}), so they don't show a full picture of their performance. For example, a model that does not perform that well in the \textit{Asian Females} demographic group, if evaluated on a test set which barely contains samples from this group, will see little or no effect on its overall performance and will appear to be a good model.

The relatively low performance in some groups seems to be originated by a limited ability to capture the best discriminant features for the samples underrepresented in the training databases. ResNet-50 seems to learn better discriminant features as it performs better than VGG-Face. Additionally, ResNet-50 shows smaller difference between demographic groups. The results suggest that features capable of reaching high accuracy for a specific demographic group may be less competitive in others.

\begin{figure*}
    \centering
         \subfloat[DiveFace ($\textbf{x}$)]{\includegraphics[width=0.3\textwidth]{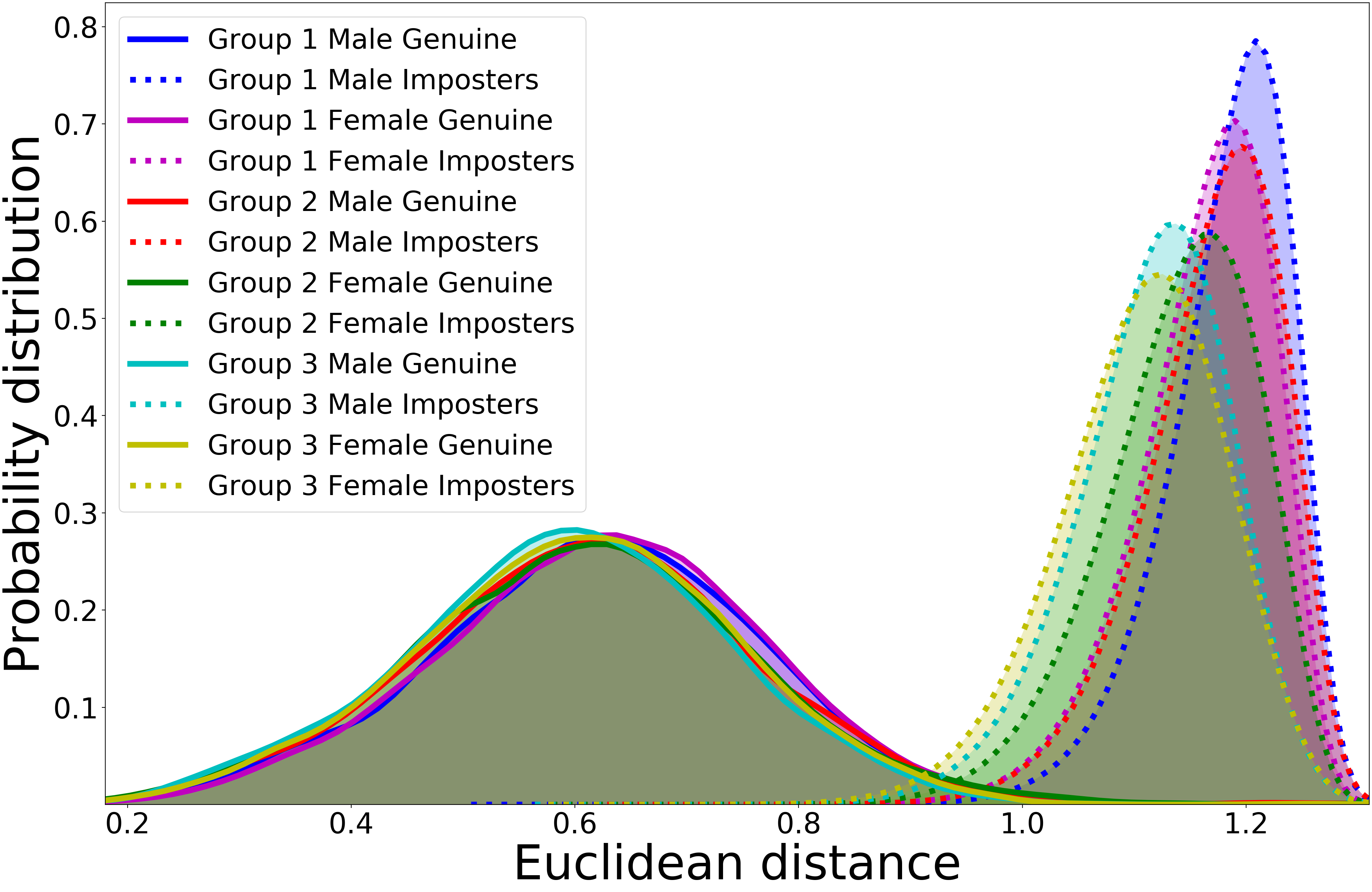}
         \label{fig:ResNet_DiveFace}}
    \hfil
         \subfloat[RFW ($\textbf{x}$)]{\includegraphics[width=0.3\textwidth]{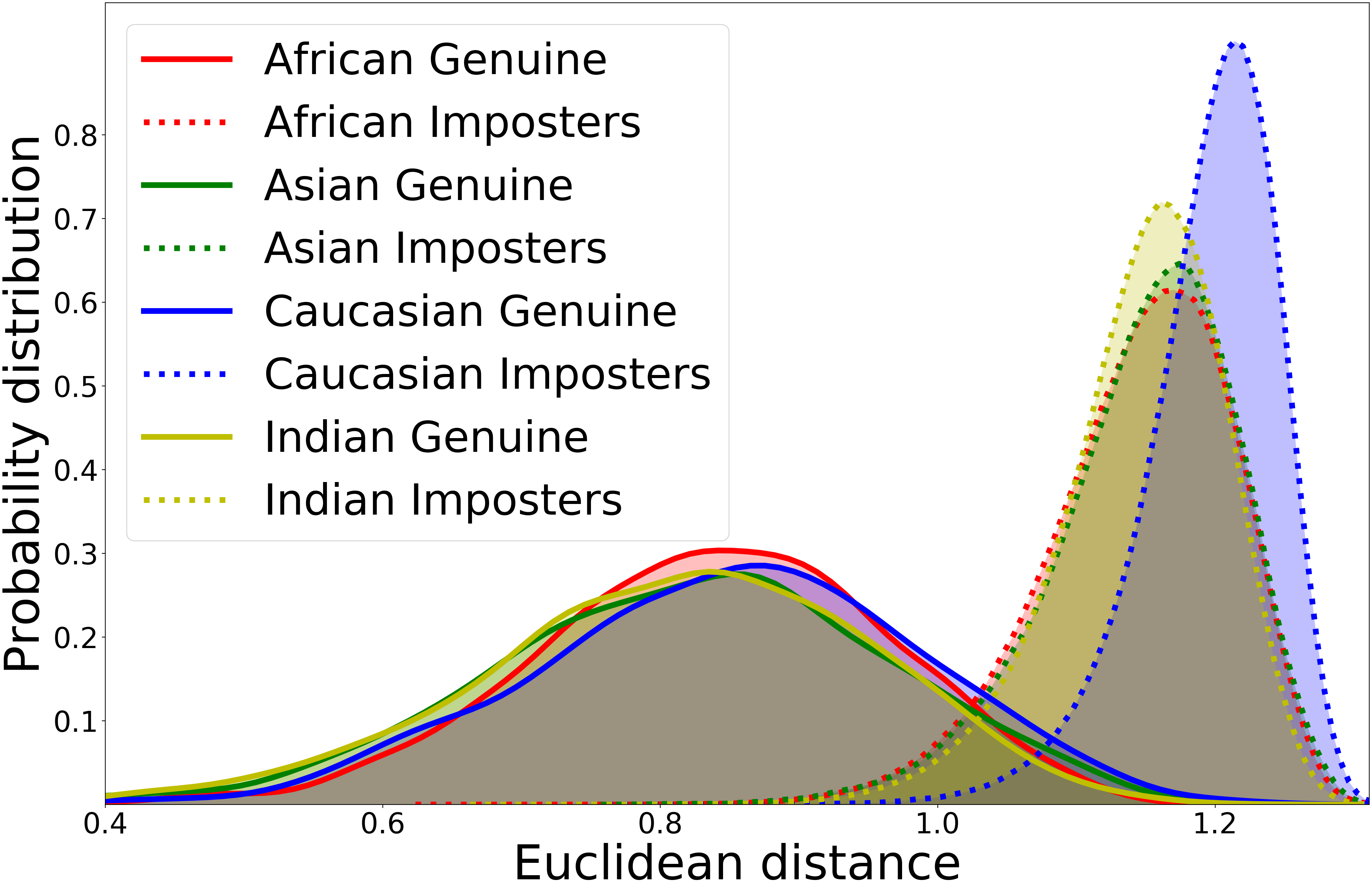}
         \label{fig:ResNet_RFW}}
    \hfil
         \subfloat[BUPT-B ($\textbf{x}$)]{\includegraphics[width=0.3\textwidth]{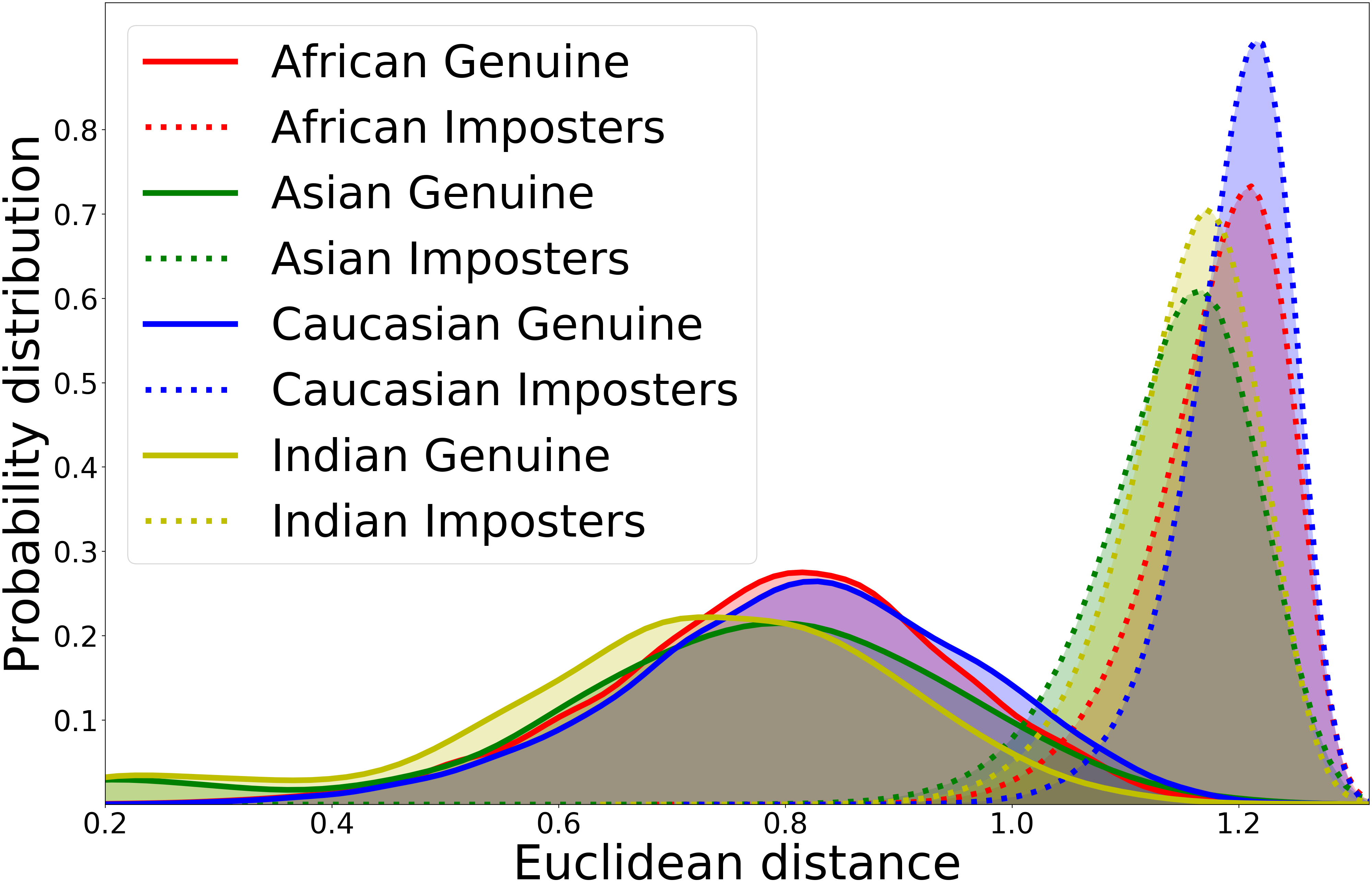}
         \label{fig:ResNet_BUPT}}
    \hfil
         \subfloat[DiveFace ($\bm{\upvarphi}(\textbf{x})$)]{\includegraphics[width=0.3\textwidth]{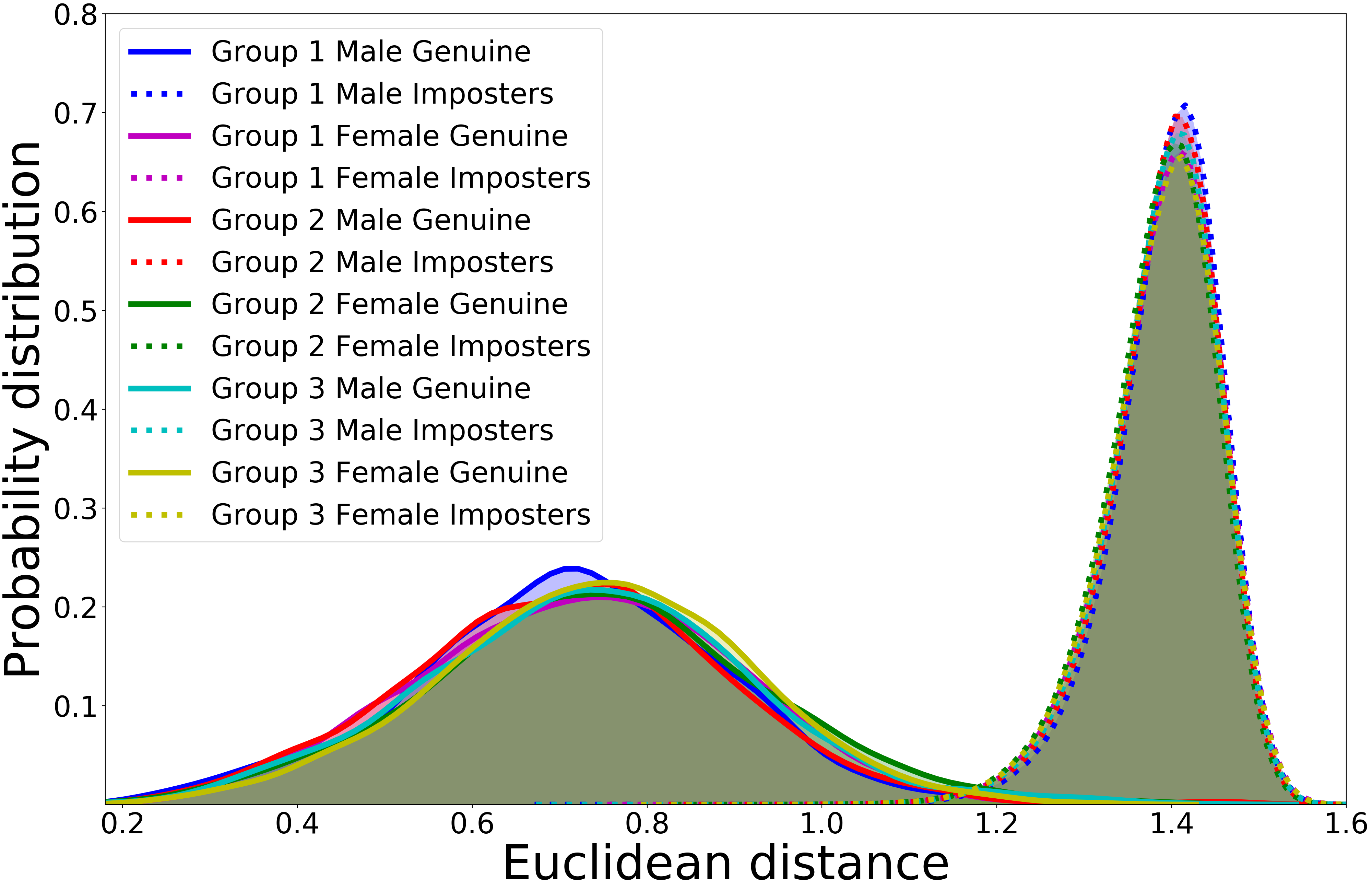}
         \label{fig:Module_DiveFace}}
    \hfil
         \subfloat[RFW ($\bm{\upvarphi}(\textbf{x})$)]{\includegraphics[width=0.3\textwidth]{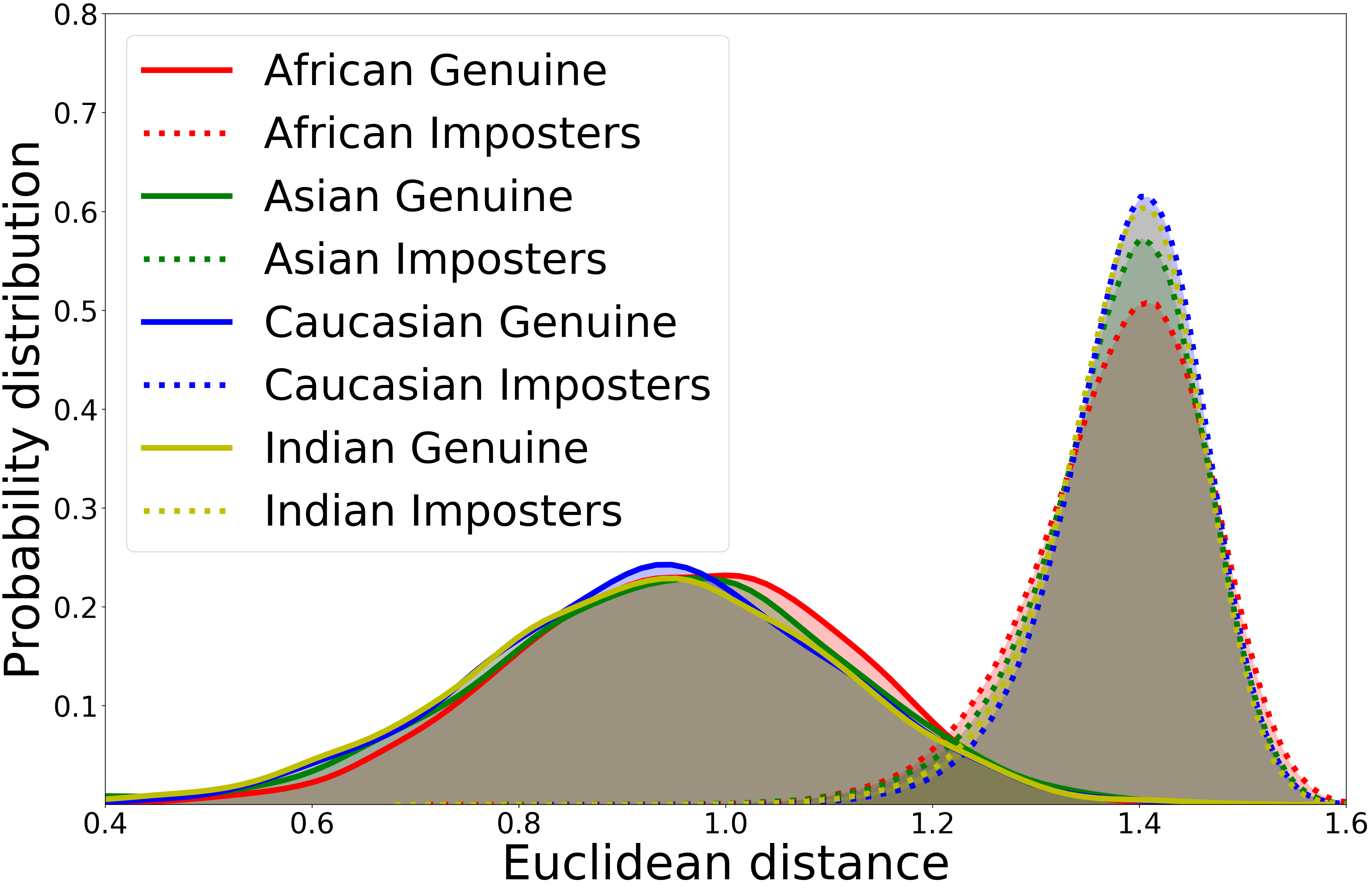}
         \label{fig:Module_RFW}}
    \hfil
         \subfloat[BUPT-B ($\bm{\upvarphi}(\textbf{x})$)]{\includegraphics[width=0.3\textwidth]{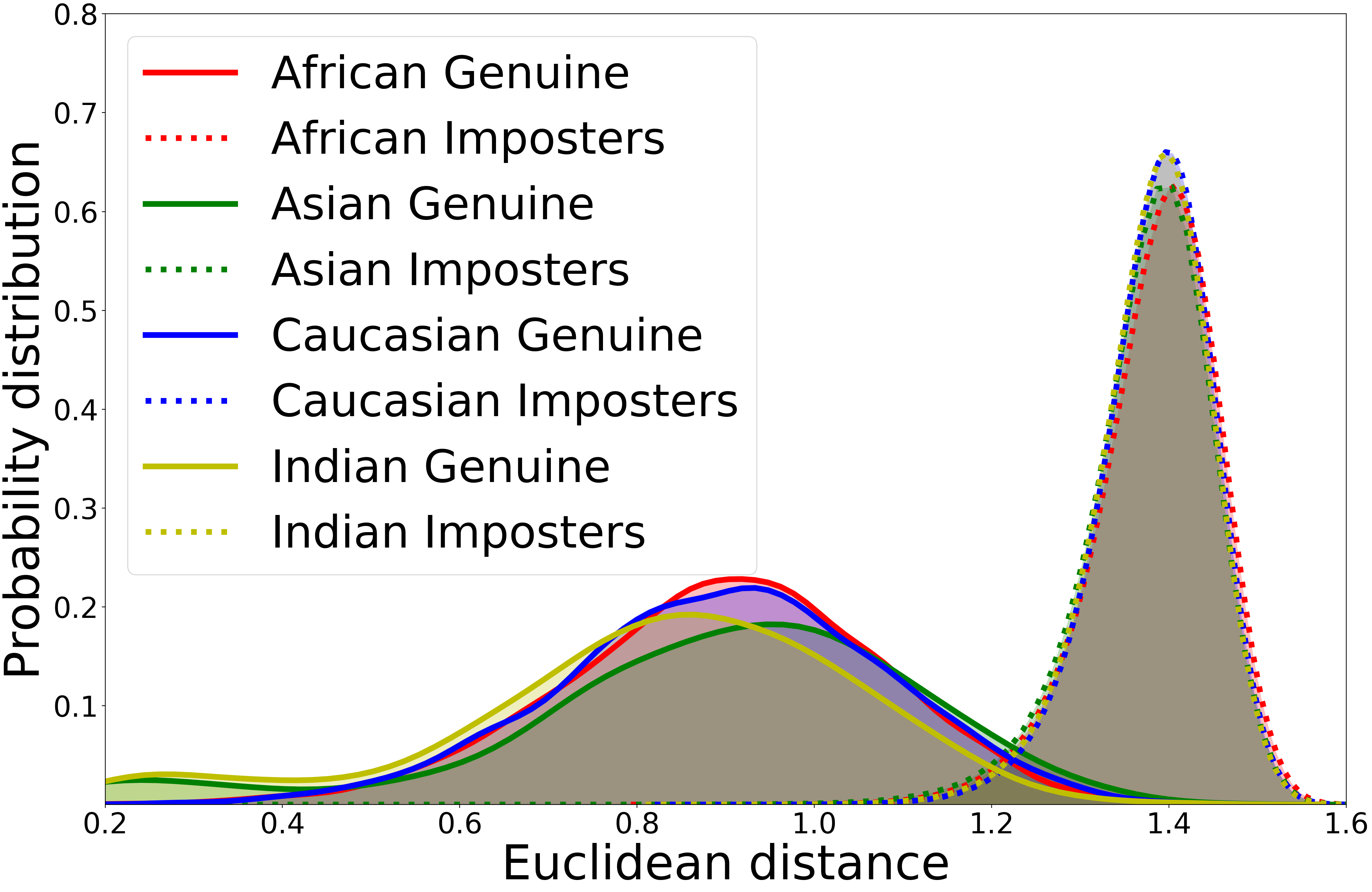}
         \label{fig:Module_BUPT}}
    \caption{ResNet-50 face verification distance score distributions for all DiveFace, RFW and BUPT-B demographic groups using the original representation $\textbf{x}$ (top) and the proposed representation $\bm{\upvarphi}(\textbf{x})$ (bottom). Note how the proposed Sensitive Loss representation $\bm{\upvarphi}(\textbf{x})$ reduces the gap between impostor distributions (dotted lines) across demographic groups.}
    \label{distributions}
\end{figure*}

Let’s now analyze the causes behind this degradation. Fig. \ref{distributions} represents the probability distributions of genuine and impostor distance scores for all demographic groups. A comparison between genuine and impostor distributions reveals large differences for impostors. The genuine distribution (intra-class variability) between groups is similar, but the impostor distribution (inter-class variability) is significantly different. The baseline models behave differently between demographic groups when comparing face features from different people.

Fig. \ref{distributions} shows the score distributions obtained for the ResNet-50 model without and with our Sensitive Loss de-biasing method (with Unrestricted sensitive triplet generation). Table \ref{tabla_EER} showed performances for specific decision thresholds (at the EER) for face verification. Fig. \ref{distributions} provides richer information without fixing the decision thresholds. In comparison to the baseline $\textbf{x}$, we see that the improvements in Accuracy and Fairness caused by our Sensitive Loss discrimination-aware representation $\bm{\upvarphi}(\textbf{x})$ mainly come from better alignment of impostor score distributions across demographic groups. These results suggest how the proposed Sensitive Loss learning method was able to correct the biased behavior of the baseline model.

The results obtained by Sensitive Loss outperform the baseline approaches by:
\begin{enumerate}
\renewcommand{\labelenumi}{\roman{enumi}}
    \item[i)] Improving the fairness (Std) with lower standard deviation in performance across demographic groups. Fairness improvements in terms of EER Std vary by model and database ranging from $5\%$ to $61\%$ relative improvements with an average improvement of $44\%$.
    \item[ii)] Reducing the Average EER in the three databases. The results show that discrimination-aware learning not only helps to train fairer representations but also more accurate ones. Our Sensitive Loss discrimination-aware learning results in better representations for specific demographic groups and collectively for all groups.
\end{enumerate}

\begin{itemize}
    \item[] \textbf{Algorithmic Discrimination implications:} define the performance function $f$ as the accuracy of the face recognition model, and $\textit{G}(\mathcal{D}_d^k)=f(\mathcal{D}_d^k,\textbf{w}^*)$ the goodness considering all the samples corresponding to class $k$ of the demographic criterion $d$, for an algorithm $\textbf{w}^*$ trained on the full set of data $\mathcal{D}$ (as described in Eq. \ref{eqn:learning_strategy}). Results suggest large differences between the goodness $\textit{G}(\mathcal{D}_d^k)$ for different classes, especially between the classes $k=\textit{Caucasian}$ and $\textit{Asian}$.
\end{itemize}

\subsubsection{Comparison with the state of the art}

Table \ref{tabla_papers} shows the comparison of our approach with two recent state-of-the-art de-biasing techniques \cite{wang2019mitigate, gong2019debface}. These two methods consist of full networks trained specifically to avoid bias, whereas what we propose here with Sensitive Loss is not an entire network, but rather an add-on method to reduce the biased outcome of a given network. 

The results of this comparison should be interpreted with care, because the arrangements are different. Still, the comparison gives us a rough idea of the ranges of bias mitigation in the three methods. 

The two approaches we compared ourselves with, have trained the network with only one database: BUPT-BalanceFace. We have instead taken a network already trained with VGGFace2 and added a layer that we have trained with BUPT-BalanceFace. Our network may have an advantage because one part has been trained with VGGFace2 and the other with BUPT-BalanceFace and therefore the average performance is better. However, we are not looking to improve performance, but to reduce discrimination, and with our experiments we want to demonstrate that complex models are not always needed. 


DebFace or RL-RBN cannot be compared to our ArcFace-based method because the RFW database is included in MS1M (ArcFace training data set). In fact, both the EER and Std of ArcFace in RFW is up to 10 times lower than that achieved with these. That is why we used the ResNet-50 network for the comparison. We also tested our method using the softmax loss function instead of the triplet loss function. It can be seen that remarkable results are also achieved, although not as good as with triplet loss.

\begin{table*}
\normalsize
  \begin{center}
    \caption{Comparison with state-of-the-art de-biasing approaches. EER in \% in a face verification task on the RFW database. In brackets we show the relative improvement with respect to the baseline approach used in each work. NA = Not Available.}\smallskip
    \label{tabla_papers}
    \begin{tabular}{l|c|c|c|c||c|c} 
      \toprule
      \textbf{Method} & \textbf{Caucasian} & \textbf{Indian} & \textbf{African} & \textbf{Asian} & \textbf{Avg} & \textbf{Std}\\
      \midrule 
      \Tspace
      RL-RBN (arc) \cite{wang2019mitigate} & 3.73 &	5.32 &	5 &	5.18 & $4.81$ (\footnotesize$10\%$) & $0.63$ (\footnotesize 34\%)\\ 
      \midrule 
      \Tspace
      DebFace \cite{gong2019debface} & 4.05 &	5.22 &	6.33 &	5.67 & $5.32$ (\footnotesize NA) & $0.83$ (\footnotesize NA)\\
      \midrule 
      \Tspace
      Sensitive Loss & 2.77 &	3.05 &	4.18 &	3.50 & $3.37$ (\footnotesize $33\%$) & $0.54$ (\footnotesize $42\%$)\\ 
      (Proposed Here) & 3.17 &	3.38 &	4.69 &	3.84 & $3.77$ (\footnotesize $25\%$) & $0.58$ (\footnotesize $38\%$)\\ 
      \bottomrule 
    \end{tabular}
  \end{center}
\end{table*}

From Table \ref{tabla_papers} it can be seen that in terms of fairness (measured as performance differences among demographic groups) our approach is at least comparable to that of dedicated networks trained from scratch to produce unbiased models. Given that similar behavior in terms of fairness improvement, our proposed Sensitive Loss is superior to the compared methods in simplicity and applicability, as it can be directly applied as an add-on to already trained networks without the need for complete retraining.

\begin{itemize}
    \item[]\textbf{Algorithmic Discrimination implications:} the discrimination-aware learning method proposed in this work, Sensitive Loss, is a step forward to prevent discriminatory effects in the usage of automatic face recognition systems. The representation $\bm{\upvarphi}(\textbf{x})$ reduces the discriminatory effects of the original representation $\textbf{x}$ as differences between goodness criteria $\textit{G}(\mathcal{D}_d^k)$ across demographic groups are reduced. However, differences still exist and should be considered in the deployment of these technologies. 
\end{itemize}

\section{Conclusions}

We have presented a comprehensive analysis of face recognition models based on deep learning according to a new discrimination-aware perspective. We started presenting a new general formulation of Algorithmic Discrimination with application to face recognition. We then showed the high bias introduced when training the deep models with the most popular face databases employed in the literature. We then evaluated three popular pre-trained face models (VGG-Face, ResNet-50 and ArcFace) according to the proposed formulation. 

The experiments are carried out on three public databases (DiveFace, RFW, and BUPT-B) comprising 64,000 identities and 1.5M images. The results show that the two tested face models are highly biased across demographic groups. In particular, we observed large performance differences in face recognition across gender and ethnic groups. These performance gaps reached up to 200\% of relative error degradation between the best class and the worst. This means that false positives are 200\% more likely for some demographic groups than for others when using the popular face models evaluated in this work.

We also looked at the interior of the tested models, revealing different activation patterns of the networks for different demographic groups. This corroborates the biased nature of these popular pre-trained face models.

After the bias analysis, we proposed a novel discrimination-aware training method, Sensitive Loss, based on a triplet loss function and online selection of sensitive triplets. Different to related existing de-biasing methods, Sensitive Loss works as an add-on to pre-trained networks, thereby facilitating its application to problems (like face recognition) where hard-worked models exist with excellent performance, but little attention about fairness aspects were considered in their inception. Experiments with Sensitive Loss demonstrate how simple discrimination-aware rules can guide the learning process towards fairer and more accurate representations. The results of the proposed Sensitive Loss representation outperform the baseline models for the three evaluated databases both in terms of average accuracy and fairness metrics. These results encourage the training of more diverse models and the development of methods capable of dealing with the differences inherent to demographic groups. 

The framework analyzed in this work is focused on the analysis of Group-based Algorithmic Discrimination (G-AD). Future work will investigate how to incorporate User-specific Algorithmic Discrimination (U-AD) in the proposed framework. Additionally, the analysis of other covariates such as the age will be included in the study. Discrimination by age is an important concern in applications such as automatic recruitment tools. Other future directions include the study of new methods to detect bias in the training process in an unsupervised way or the application of privacy-preserving techniques at image level \cite{Ross2019privacy}.

\section{Acknowledgments}

This work has been supported by projects: PRIMA (MSCA-ITN-2019-860315), TRESPASS-ETN (MSCA-ITN-2019-860813), BIBECA (RTI2018-101248-B-I00 MINECO/FEDER), and Accenture. I. Serna is supported by a research fellowship from the Spanish CAM (PEJD-2018-PRE/TIC-9449).

\ifCLASSOPTIONcaptionsoff
  \newpage
\fi



\bibliographystyle{IEEEtran}
\bibliography{IEEE}

\begin{thebibliography}{10}
\providecommand{\url}[1]{#1}
\csname url@samestyle\endcsname
\providecommand{\newblock}{\relax}
\providecommand{\bibinfo}[2]{#2}
\providecommand{\BIBentrySTDinterwordspacing}{\spaceskip=0pt\relax}
\providecommand{\BIBentryALTinterwordstretchfactor}{4}
\providecommand{\BIBentryALTinterwordspacing}{\spaceskip=\fontdimen2\font plus
\BIBentryALTinterwordstretchfactor\fontdimen3\font minus
  \fontdimen4\font\relax}
\providecommand{\BIBforeignlanguage}[2]{{%
\expandafter\ifx\csname l@#1\endcsname\relax
\typeout{** WARNING: IEEEtran.bst: No hyphenation pattern has been}%
\typeout{** loaded for the language `#1'. Using the pattern for}%
\typeout{** the default language instead.}%
\else
\language=\csname l@#1\endcsname
\fi
#2}}
\providecommand{\BIBdecl}{\relax}
\BIBdecl

\bibitem{rahwan2019machine}
I.~Rahwan, M.~Cebrian, N.~Obradovich \emph{et~al.}, ``{Machine Behaviour},''
  \emph{Nature}, vol. 568, no. 7753, pp. 477--486, 2019.

\bibitem{russell2016AI}
S.~Russell and P.~Norvig, \emph{{Artificial Intelligence: A Modern
  Approach}}.\hskip 1em plus 0.5em minus 0.4em\relax Pearson, 2016.

\bibitem{Patel2018survey}
R.~{Ranjan}, S.~{Sankaranarayanan}, A.~{Bansal}, N.~{Bodla}, J.~{Chen}, V.~M.
  {Patel}, C.~D. {Castillo}, and R.~{Chellappa}, ``{Deep Learning for
  Understanding Faces: Machines May Be Just as Good, or Better, than Humans},''
  \emph{IEEE Signal Processing Magazine}, vol.~35, no.~1, pp. 66--83, 2018.

\bibitem{marcel2020biometrics}
X.~Akhtar, A.~Hadid, M.~Nixon, M.~Tistarelli, J.~Dugelay, and S.~Marcel,
  ``{Biometrics: In search of identity and security (Q \& A)},'' \emph{IEEE
  MultiMedia}, vol.~25, no.~3, pp. 22--35, 2018.

\bibitem{Kumar2017book}
B.~Bhanu and A.~Kumar, \emph{{Deep Learning for Biometrics}}, ser. Advances in
  Computer Vision and Pattern Recognition (ACVPR).\hskip 1em plus 0.5em minus
  0.4em\relax Springer, 2017.

\bibitem{hospedales2020deep}
L.~Shao, P.~Hubert, and T.~Hospedales, ``{Special Issue on Machine Vision with
  Deep Learning},'' \emph{International Journal of Computer Vision}, vol. 128,
  p. 771–772, 2020.

\bibitem{grother2018FRVT}
P.~J. Grother, M.~L. Ngan, and K.~K. Hanaoka, \emph{{Ongoing Face Recognition
  Vendor Test (FRVT) Part 2: Identification}}, ser. NIST Internal Report.\hskip
  1em plus 0.5em minus 0.4em\relax National Institute of Standards and
  Technology, 2018.

\bibitem{cook2019demographic}
C.~M. Cook, J.~J. Howard, Y.~B. Sirotin, J.~L. Tipton, and A.~R. Vemury,
  ``{Demographic Effects in Facial Recognition and Their Dependence on Image
  Acquisition: An Evaluation of Eleven Commercial Systems},'' \emph{IEEE
  Transactions on Biometrics, Behavior, and Identity Science}, vol.~1, no.~1,
  pp. 32--41, 2019.

\bibitem{chellapa2019face}
B.~Lu, J.-C. Chen, C.~D. Castillo, and R.~Chellappa, ``{An experimental
  Evaluation of Covariates Effects on Unconstrained Face Verification},''
  \emph{IEEE Transactions on Biometrics, Behavior, and Identity Science},
  vol.~1, no.~1, pp. 42--55, 2019.

\bibitem{acien2018bias}
A.~Acien, A.~Morales, R.~Vera-Rodriguez, I.~Bartolome, and J.~Fierrez,
  ``{Measuring the Gender and Ethnicity Bias in Deep Models for Face
  Recognition},'' in \emph{Iberoamerican Congress on Pattern
  Recognition}.\hskip 1em plus 0.5em minus 0.4em\relax Madrid, Spain: Springer,
  2018, pp. 584--593.

\bibitem{Bowyer2020face}
K.~Krishnapriya, V.~Albiero, K.~Vangara, M.~King, and K.~Bowyer, ``{Issues
  Related to Face Recognition Accuracy Varying Based on Race and Skin Tone},''
  \emph{IEEE Transactions on Technology and Society}, vol.~1, pp. 8--20, 2020.

\bibitem{klare2012demographic}
B.~F. Klare, M.~J. Burge, J.~C. Klontz, R.~W.~V. Bruegge, and A.~K. Jain,
  ``{Face Recognition Performance: Role of Demographic Information},''
  \emph{IEEE Transactions on Information Forensics and Security}, vol.~7,
  no.~6, pp. 1789--1801, 2012.

\bibitem{buolamwini2018GenderShades}
J.~Buolamwini and T.~Gebru, ``{Gender Shades: Intersectional Accuracy
  Disparities in Commercial Gender Classification},'' in \emph{Conference on
  Fairness, Accountability and Transparency}, ser. Proceedings of Machine
  Learning Research, S.~A. Friedler and C.~Wilson, Eds., vol.~81, New York, NY,
  USA, 23--24 Feb 2018, pp. 77--91.

\bibitem{zisserman2018BlindEye}
M.~Alvi, A.~Zisserman, and C.~Nell{\aa}ker, ``{Turning a Blind Eye: Explicit
  Removal of Biases and Variation from Deep Neural Network embeddings},'' in
  \emph{European Conference on Computer Vision (ECCV)}, Munich, Germany, 2018,
  pp. 556--572.

\bibitem{isabelle2019DemogPairs}
I.~Hupont and C.~Fernandez, ``{DemogPairs: Quantifying the Impact of
  Demographic Imbalance in Deep Face Recognition},'' in \emph{International
  Conference on Automatic Face \& Gesture Recognition (FG)}, Lille, France,
  2019, pp. 1--7.

\bibitem{drozdowski2020bias}
P.~{Drozdowski}, C.~{Rathgeb}, A.~{Dantcheva}, N.~{Damer}, and C.~{Busch},
  ``{Demographic Bias in Biometrics: A Survey on an Emerging Challenge},''
  \emph{IEEE Transactions on Technology and Society}, vol.~1, no.~2, pp.
  89--103, 2020.

\bibitem{wang2019mitigate}
M.~Wang and W.~Deng, ``{Mitigate Bias in Face Recognition using Skewness-Aware
  Reinforcement Learning},'' in \emph{Conference on Computer Vision and Pattern
  Recognition (CVPR)}.\hskip 1em plus 0.5em minus 0.4em\relax Seattle,
  Washington, USA: IEEE, 2020, pp. 9319--9328.

\bibitem{gong2019debface}
S.~Gong, X.~Liu, and A.~Jain, ``{Jointly De-biasing Face Recognition and
  Demographic Attribute Estimation},'' in \emph{European Conference on Computer
  Vision}, Virtual, August 2020.

\bibitem{serna2020discrimination}
I.~Serna, A.~Morales, J.~Fierrez, N.~Cebrian, M.~Obradovich, and I.~Rahwan,
  ``{Algorithmic Discrimination: Formulation and Exploration in Deep
  Learning-based Face Biometrics},'' in \emph{AAAI Workshop on Artificial
  Intelligence Safety (SafeAI)}, New York, NY, USA, 2020.

\bibitem{kleinberg2019discrimination}
J.~Kleinberg, J.~Ludwig, S.~Mullainathan, and C.~R. Sunstein, ``{Discrimination
  in the Age of Algorithms},'' \emph{Journal of Legal Analysis}, vol.~10, pp.
  113--174, 04 2019.

\bibitem{ranjan2018faces}
R.~Ranjan, S.~Sankaranarayanan \emph{et~al.}, ``{Deep Learning for
  Understanding Faces: Machines May Be Just as Good, or Better, than Humans},''
  \emph{IEEE Signal Processing Magazine}, vol.~35, no.~1, pp. 66--83, 2018.

\bibitem{he2016resnet}
K.~He, X.~Zhang, S.~Ren, and J.~Sun, ``{Deep Residual Learning for Image
  Recognition},'' in \emph{Conference on Computer Vision and Pattern
  Recognition (CVPR)}.\hskip 1em plus 0.5em minus 0.4em\relax Las Vegas, NV,
  USA: IEEE, 2016, pp. 770--778.

\bibitem{ratha2019diversity}
M.~Merler, N.~Ratha, R.~S. Feris, and J.~R. Smith, ``{Diversity in Faces},''
  \emph{arXiv:1901.10436}, pp. 1--29, 2019.

\bibitem{grother2019FRVT}
P.~J. Grother, M.~L. Ngan, and K.~K. Hanaoka, \emph{{Ongoing Face Recognition
  Vendor Test (FRVT) Part 3: Demographic Effects}}, ser. NIST Internal
  Report.\hskip 1em plus 0.5em minus 0.4em\relax U.S. Department of Commerce,
  National Institute of Standards and Technology, 2019.

\bibitem{2016msceleb}
Y.~Guo, L.~Zhang, Y.~Hu, X.~He, and J.~Gao, ``{Ms-celeb-1m: A Dataset and
  Benchmark for Large-Scale Face Recognition},'' in \emph{European Conference
  on Computer Vision (ECCV)}.\hskip 1em plus 0.5em minus 0.4em\relax Amsterdam,
  The Netherlands: Springer, 2016, pp. 87--102.

\bibitem{2016Megaface}
I.~Kemelmacher-Shlizerman, S.~M. Seitz, D.~Miller, and E.~Brossard, ``{The
  Megaface Benchmark: 1 Million Faces for Recognition at Scale},'' in
  \emph{Conference on Computer Vision and Pattern Recognition (CVPR)}.\hskip
  1em plus 0.5em minus 0.4em\relax Las Vegas, Nevada, USA: IEEE, 2016, pp.
  4873--4882.

\bibitem{cao2018vgg2}
Q.~Cao, L.~Shen, W.~Xie, O.~M. Parkhi, and A.~Zisserman, ``{Vggface2: A Dataset
  for Recognising Faces Across Pose and Age},'' in \emph{International
  Conference on Automatic Face \& Gesture Recognition (FG)}.\hskip 1em plus
  0.5em minus 0.4em\relax Lille, France: IEEE, 2018, pp. 67--74.

\bibitem{parkhi2015face}
O.~M. Parkhi, A.~Vedaldi, A.~Zisserman \emph{et~al.}, ``{Deep Face
  Recognition},'' in \emph{British Machine Vision Conference (BMVC)}, Swansea,
  UK, 2015, pp. 41.1--41.12.

\bibitem{2011youtubedb}
L.~Wolf, T.~Hassner, and I.~Maoz, ``{Face Recognition in Unconstrained Videos
  with Matched Background Similarity},'' in \emph{Computer Vision and Pattern
  Recognition (CVPR)}.\hskip 1em plus 0.5em minus 0.4em\relax Colorado Springs,
  CO, USA: IEEE, June 2011, pp. 529--534.

\bibitem{2014Casiadb}
D.~Yi, Z.~Lei, S.~Liao, and S.~Z. Li, ``{Learning Face Representation from
  Scratch},'' \emph{arXiv:1411.7923}, pp. 1--9, 2014.

\bibitem{2015CelebA}
S.~Yang, P.~Luo, C.-C. Loy, and X.~Tang, ``{From Facial Parts Responses to Face
  Detection: A Deep Learning Approach},'' in \emph{International Conference on
  Computer Vision (ICCV)}, Santiago, Chile, 2015, pp. 3676--3684.

\bibitem{kumar2011face}
N.~Kumar, A.~Berg, P.~N. Belhumeur, and S.~Nayar, ``{Describable Visual
  Attributes for Face Verification and Image Search},'' \emph{IEEE Transactions
  on Pattern Analysis and Machine Intelligence}, vol.~33, no.~10, pp.
  1962--1977, 2011.

\bibitem{2018IJBc}
B.~Maze, J.~Adams, J.~A. Duncan, N.~Kalka, T.~Miller, C.~Otto, A.~K. Jain,
  W.~T. Niggel, J.~Anderson, J.~Cheney \emph{et~al.}, ``{IARPA Janus
  Benchmark-C: Face Dataset and Protocol},'' in \emph{International Conference
  on Biometrics (ICB)}.\hskip 1em plus 0.5em minus 0.4em\relax Gold Coast,
  Australia: IEEE, 2018, pp. 158--165.

\bibitem{utk2017age}
Z.~Zhang, Y.~Song, and H.~Qi, ``{Age Progression/Regression by Conditional
  Adversarial Autoencoder},'' in \emph{Conference on Computer Vision and
  Pattern Recognition (CVPR)}.\hskip 1em plus 0.5em minus 0.4em\relax Honolulu,
  Hawaii, USA: IEEE, 2017, pp. 5810--5818.

\bibitem{miller2007LFW}
G.~B. Huang, M.~Ramesh, T.~Berg, and E.~Learned-Miller, ``{Labeled Faces in the
  Wild: A Database for Studying Face Recognition in Unconstrained
  Environments},'' University of Massachusetts, Amherst, Tech. Rep. 07-49,
  October 2007.

\bibitem{ortega2009BMDB}
J.~Ortega-Garcia, J.~Fierrez \emph{et~al.}, ``{The Multiscenario
  Multienvironment Biosecure Multimodal Database (BMDB)},'' \emph{IEEE
  Transactions on Pattern Analysis and Machine Intelligence}, vol.~32, no.~6,
  pp. 1097--1111, 2009.

\bibitem{SesitiveNets2019}
A.~Morales, J.~Fierrez, R.~Vera-Rodriguez, and R.~Tolosana, ``{SensitiveNets:
  Learning Agnostic Representations with Application to Face Recognition},''
  \emph{IEEE Transactions on Pattern Analysis and Machine Intelligence}, pp.
  1--8, 2020.

\bibitem{krkkinen2019fairface}
K.~Kärkkäinen and J.~Joo, ``{FairFace: Face Attribute Dataset for Balanced
  Race, Gender, and Age},'' \emph{arXiv:1908.04913}, pp. 1--11, 2019.

\bibitem{wang2019RFW}
M.~Wang, W.~Deng, J.~Hu, X.~Tao, and Y.~Huang, ``{Racial Faces in the Wild:
  Reducing Racial Bias by Information Maximization Adaptation Network},'' in
  \emph{International Conference on Computer Vision (ICCV)}.\hskip 1em plus
  0.5em minus 0.4em\relax Seoul, Korea: IEEE, October 2019, pp. 692--702.

\bibitem{das2018mitigatebias}
A.~Das, A.~Dantcheva, and F.~Bremond, ``{Mitigating Bias in Gender, Age and
  Ethnicity Classification: a Multi-Task Convolution Neural Network
  Approach},'' in \emph{European Conference on Computer Vision (ECCV)}, Munich,
  Germany, 2018, pp. 573--585.

\bibitem{calders2010discrimination}
T.~Calders and S.~Verwer, ``{Three Naive Bayes Approaches for
  Discrimination-Free Classification},'' \emph{Data Mining and Knowledge
  Discovery}, vol.~21, no.~2, pp. 277--292, 2010.

\bibitem{raji2019bias}
I.~D. Raji and J.~Buolamwini, ``{Actionable Auditing: Investigating the Impact
  of Publicly Naming Biased Performance Results of Commercial AI Products},''
  in \emph{Conference on AI Ethics and Society (AIES)}.\hskip 1em plus 0.5em
  minus 0.4em\relax New York, NY, USA: AAAI/ACM, 2019, pp. 429--435.

\bibitem{fierrez2011quality}
F.~Alonso-Fernandez, J.~Fierrez, and J.~Ortega-Garcia, ``{Quality Measures in
  Biometric Systems},'' \emph{IEEE Security \& Privacy}, vol.~10, no.~6, pp.
  52--62, 2011.

\bibitem{pentland2020fair}
M.~Bakker, H.~R. Valdes, D.~P. Tu, K.~Gummadi, K.~Varshney, A.~Weller, and
  A.~Pentland, ``{Fair Enough: Improving Fairness in Budget-Constrained
  Decision Making Using Confidence Thresholds},'' in \emph{AAAI Workshop on
  Artificial Intelligence Safety (SafeAI)}, New York, NY, USA, 2020, pp.
  41--53.

\bibitem{varshney2020AIfairness}
Y.~Zhang, R.~Bellamy, and K.~Varshney, ``{Joint Optimization of AI Fairness and
  Utility: A Human-Centered Approach},'' in \emph{Conference on AI, Ethics, and
  Society (AIES)}.\hskip 1em plus 0.5em minus 0.4em\relax New York, NY, USA:
  AAAI/ACM, 2020, pp. 400--406.

\bibitem{weinberger2006Triplet}
K.~Q. Weinberger and L.~K. Saul, ``{Distance Metric Learning for Large Margin
  Nearest Neighbor Classification},'' in \emph{Advances in Neural Information
  Processing Systems (NIPS)}.\hskip 1em plus 0.5em minus 0.4em\relax MIT Press,
  2006, pp. 1473--1480.

\bibitem{schroff2015facenet}
F.~Schroff, D.~Kalenichenko, and J.~Philbin, ``{FaceNet: A Unified Embedding
  for Face Recognition and Clustering},'' in \emph{Conference on Computer
  Vision and Pattern Recognition (CVPR)}.\hskip 1em plus 0.5em minus
  0.4em\relax IEEE, June 2015, pp. 815--823.

\bibitem{deng2019arcface}
J.~Deng, J.~Guo, N.~Xue, and S.~Zafeiriou, ``{Arcface: Additive Angular Margin
  Loss for Deep Face Recognition},'' in \emph{Conference on Computer Vision and
  Pattern Recognition (CVPR)}.\hskip 1em plus 0.5em minus 0.4em\relax Long
  Beach, California, USA: IEEE, 2019, pp. 4690--4699.

\bibitem{zhang2016detection}
K.~Zhang, Z.~Zhang, Z.~Li, and Y.~Qiao, ``{Joint Face Detection and Alignment
  Using Multitask Cascaded Convolutional Networks},'' \emph{IEEE Signal
  Processing Letters}, vol.~23, no.~10, pp. 1499--1503, 2016.

\bibitem{deng2020retina}
J.~Deng, J.~Guo, E.~Ververas, I.~Kotsia, and S.~Zafeiriou, ``{RetinaFace:
  Single-Shot Multi-Level Face Localisation in the Wild},'' in \emph{Conference
  on Computer Vision and Pattern Recognition (CVPR)}.\hskip 1em plus 0.5em
  minus 0.4em\relax Seattle, Washington, USA: IEEE, June 2020, pp. 5202--5211.

\bibitem{selvaraju2017grad}
R.~R. Selvaraju, M.~Cogswell \emph{et~al.}, ``{Grad-CAM: Visual Explanations
  from Deep Networks Via Gradient-Based Localization},'' in \emph{International
  Conference on Computer Vision (CVPR)}.\hskip 1em plus 0.5em minus 0.4em\relax
  Honolulu, Hawaii, USA: IEEE, 2017, pp. 618--626.

\bibitem{bau2020understanding}
D.~Bau, J.-Y. Zhu, H.~Strobelt, A.~Lapedriza, B.~Zhou, and A.~Torralba,
  ``{Understanding the Role of Individual Units in a Deep Neural Network},''
  \emph{Proceedings of the National Academy of Sciences}, pp. 1--8, 2020.

\bibitem{Ross2019privacy}
V.~{Mirjalili}, S.~{Raschka}, and A.~{Ross}, ``{FlowSAN: Privacy-Enhancing
  Semi-Adversarial Networks to Confound Arbitrary Face-Based Gender
  Classifiers},'' \emph{IEEE Access}, vol.~7, pp. 99\,735--99\,745, 2019.

\end{thebibliography}

%

\begin{IEEEbiography}[{\includegraphics[width=1in,height=1.25in,clip,keepaspectratio]{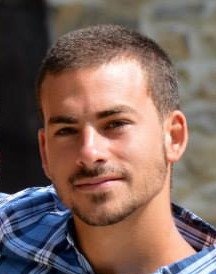}}]{Ignacio de la Serna}
recieved the B.S. degree in mathematics and the B.S. degree in computer science from the Autonomous University of Madrid, Spain, in 2018, and the M.S. degree in Artificial Intelligence from the National Distance Education University in 2020. He is currently pursuing a Ph.D. in Computer Science at the Biometrics and Data Pattern Analytics (BiDA) Lab of the School of Engineering, Autonomous University of Madrid, Spain, under the supervision of Prof. A. Morales. His research interests include computer vision, pattern recognition, and explainable AI, with applications to biometrics.
\end{IEEEbiography}

\begin{IEEEbiography}[{\includegraphics[width=1in,height=1.25in,clip,keepaspectratio]{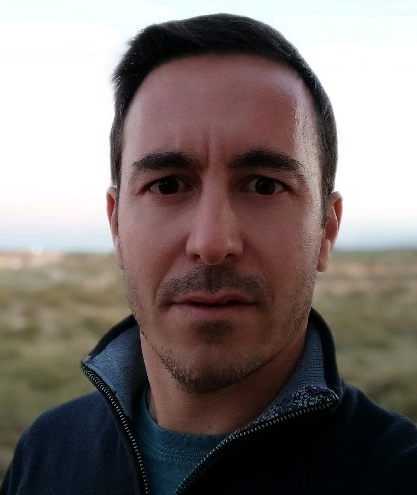}}]{Aythami Morales Moreno}
received the M.Sc. (Electronical Engineering) and Ph.D. (Artificial Intelligence) degrees from Universidad de Las Palmas de Gran Canaria in 2006 and 2011 respectively. Since 2017, he is Associate Professor with the Universidad Autonoma de Madrid. He has conducted research stays at Michigan State University, Hong Kong Polytechnic University, University of Bologna, and the Schepens Eye Research Institute. He has authored over 100 scientific articles in topics related to machine learning, trustworthy AI, and biometric signal processing.
\end{IEEEbiography}

\begin{IEEEbiography}[{\includegraphics[width=1in,height=1.25in,clip,keepaspectratio]{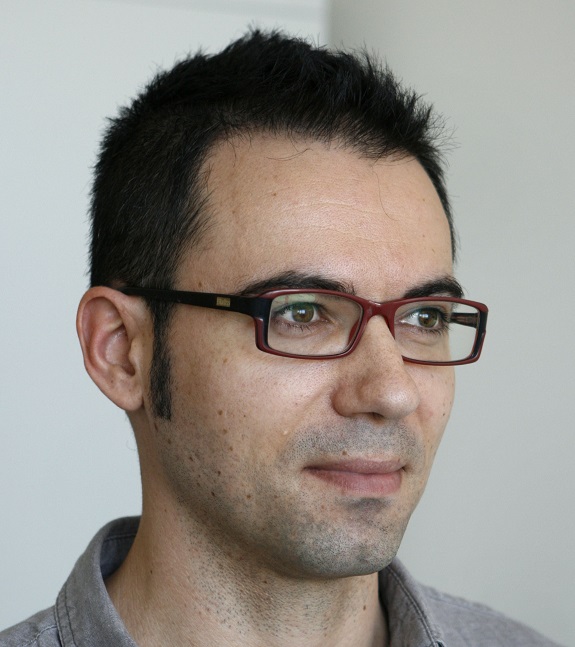}}]{Julian Fierrez}
(Member, IEEE) received the M.Sc. and Ph.D. degrees in telecommunications engineering from the Universidad Politecnica de Madrid,
Spain, in 2001 and 2006, respectively. Since 2004 he has been at Universidad Autonoma de Madrid, where he is currently an Associate
Professor. His research interests include signal and image processing, pattern recognition, and biometrics; with emphasis on evaluation, security, forensics, mobile and behavioral biometrics. He is actively involved in EU projects around biometrics (e.g., BIOSECURE, TABULA RASA and BEAT in the past; now IDEA-FAST, PRIMA, and TRESPASS-ETN). He received the Miguel Catalan Award 2015 to the Best Researcher under 40 in the Community of Madrid in the general area of science and technology, and the 2017 IAPR Young Biometrics Investigator Award. He is an Associate Editor of the IEEE TRANSACTIONS ON INFORMATION FORENSICS AND SECURITY and the IEEE TRANSACTIONS ON IMAGE PROCESSING.
\end{IEEEbiography}

\begin{IEEEbiography}[{\includegraphics[width=1in,height=1.25in,clip,keepaspectratio]{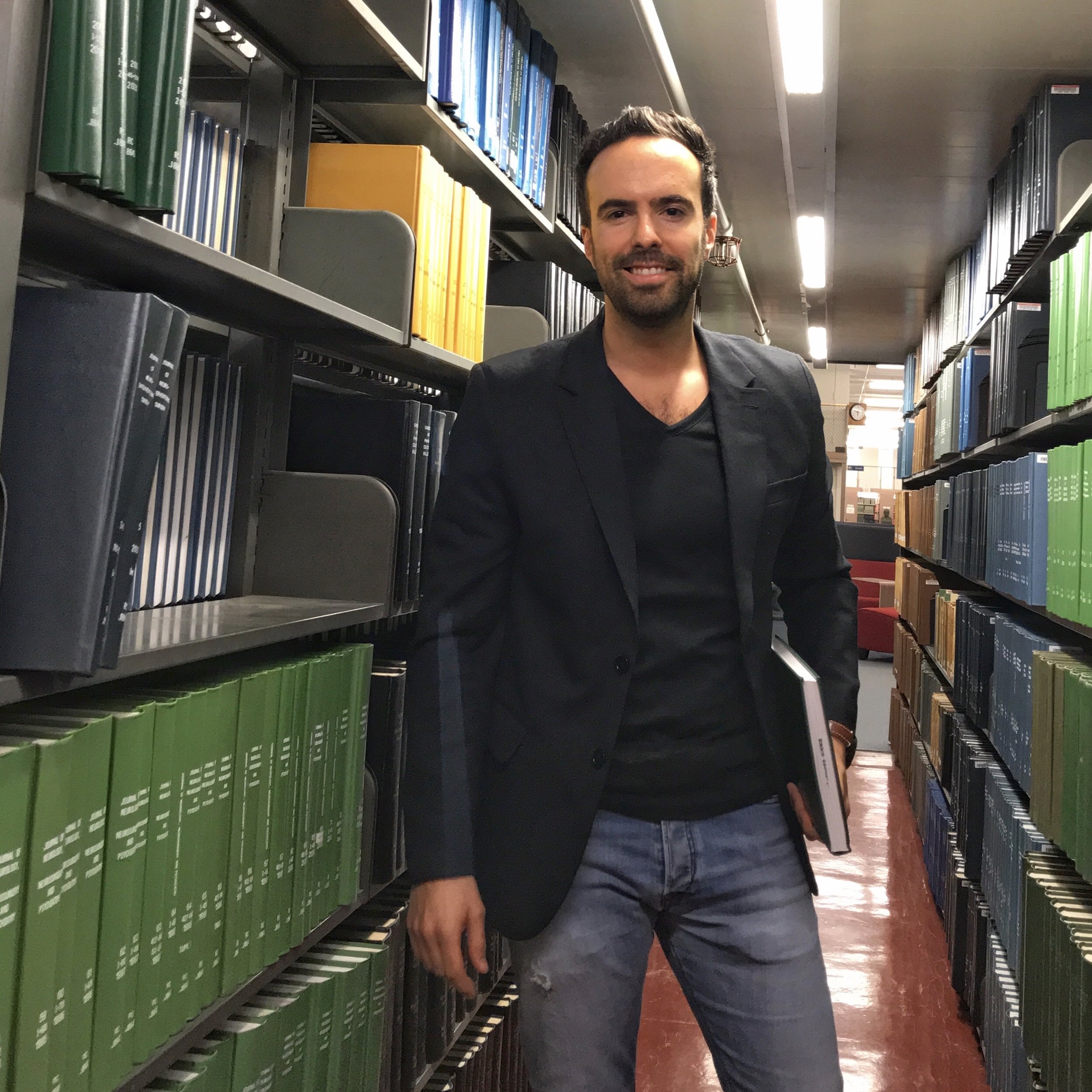}}]{Manuel Cebrian}
is a Max Planck Research Group Leader at the Max Planck Institute for Human Development. He was previously a Research Scientist Manager at the MIT Media Lab. Manuel's research examines computational methods to create incentives that mobilize large groups of people to collaborate. His empirical work uses network science modeling and observational studies. His published papers appear in Science, Nature, the Proceedings of the National Academy of Sciences of the UA, the Journal of the Royal Society Interface, and other peer-reviewed journals and proceedings in computer science and computational social science.
\end{IEEEbiography}

\begin{IEEEbiography}[{\includegraphics[width=1in,height=1.25in,clip,keepaspectratio]{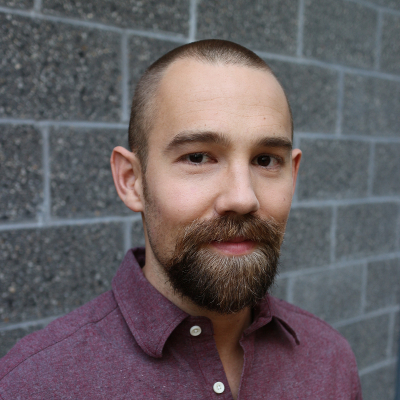}}]{Nick Obradovich}
is Senior Research Scientist and Principal Investigator at the \href{https://www.mpib-berlin.mpg.de/en/institute/max-planck-society}{Max Planck Institute for Human Development} in the \href{https://www.mpib-berlin.mpg.de/en/chm}{Center for Humans and Machines}. He holds a PhD from the University of California, San Diego and completed his postdoctoral training at Harvard University. He then worked for a number of years as a research scientist at the MIT Media Lab. In his work, he combines his interests in artificial intelligence, climate change, and human behavior with his affinity for data science and computational methods.
\end{IEEEbiography}

\begin{IEEEbiography}[{\includegraphics[width=1in,height=1.25in,clip,keepaspectratio]{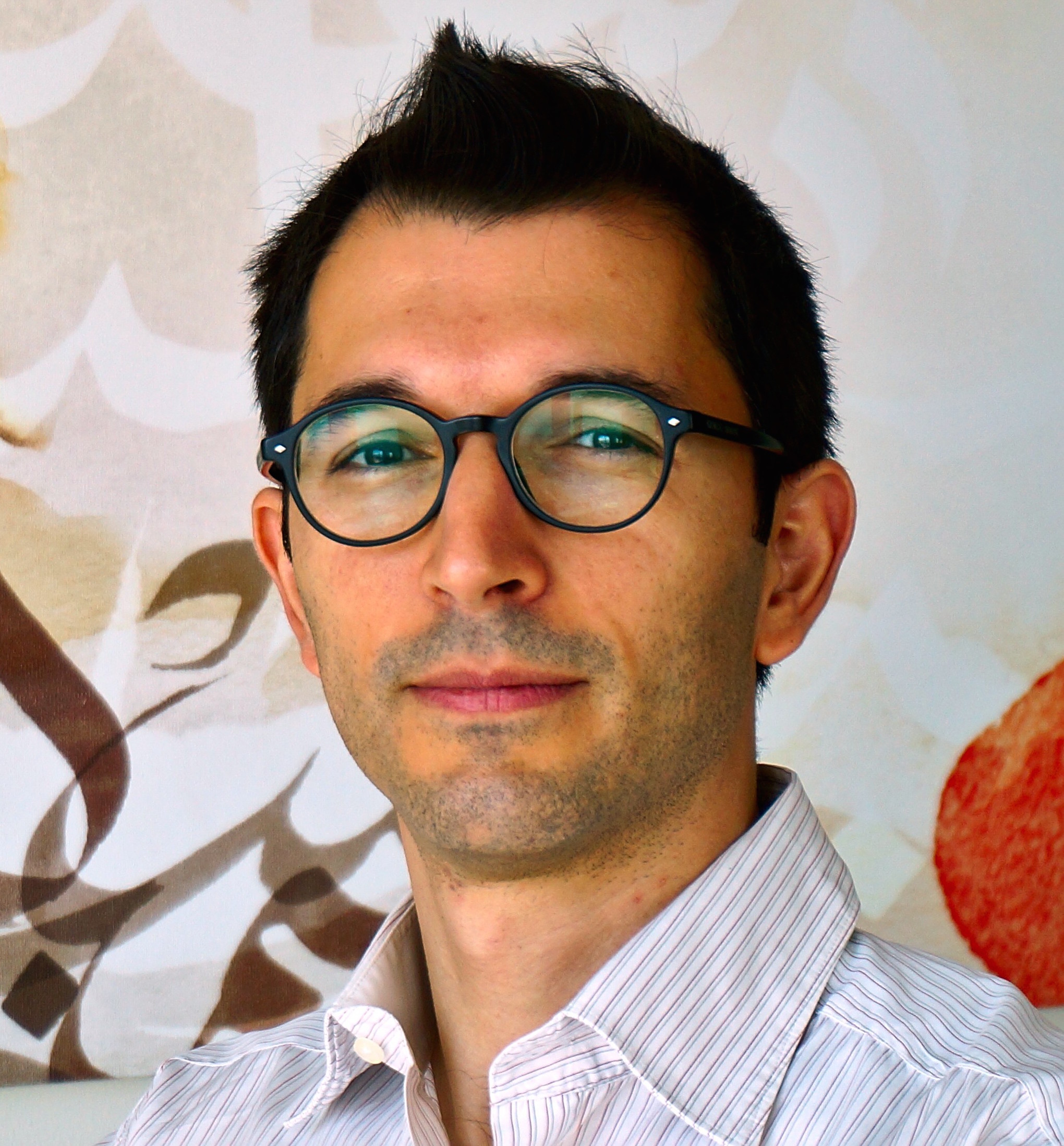}}]{Iyad Rahwan} 
is a director at the Max Planck Institute for Human Development,  where he founded the Center for Humans \& Machines. He is also an Associate Professor of Media Arts \& Sciences at the MIT Media Lab. He holds a PhD from the University of Melbourne, Australia. Rahwan's work lies at the intersection of computer science and human behavior, with a focus on collective intelligence, large-scale cooperation, and the societal impact of Artificial Intelligence and social media.
\end{IEEEbiography}






\end{document}